\PassOptionsToPackage{table,xcdraw}{xcolor}
\documentclass[letterpaper,twocolumn,10pt]{article}

\usepackage{hegroup}
\usepackage{times}
\usepackage{latexsym}
\usepackage{microtype}
\usepackage{inconsolata}
\usepackage{graphicx}
\usepackage{multirow}
\usepackage{colortbl}
\usepackage{array}
\usepackage{booktabs}
\usepackage{tabularx}
\usepackage{xcolor}
\usepackage{float}
\usepackage{booktabs}
\usepackage{amsmath}
\usepackage{cleveref}
\usepackage{adjustbox}
\usepackage{xspace}
\usepackage{enumitem}
\usepackage{subcaption}
\usepackage{makecell}
\usepackage[numbers,sort]{natbib}
\usepackage{mdframed}
\usepackage{bm}
\usepackage{caption}
\usepackage{threeparttable}
\setlist[enumerate,1]{nosep,left=0pt}
\setlist[itemize]{nosep,left=0pt}
\usepackage[most]{tcolorbox}

\newcommand{\mypara}[1]{\smallskip\noindent{\bf {#1}.}\xspace}
\newcommand{\Dataset}{\textit{SM-D}\xspace}
\newcommand{\Bench}{\textit{AIGTBench}\xspace}
\newcommand{\Detector}{\textbf{OSM-Det}\xspace}
\def\etal{\textit{et~al.}}
\newtcbox{\highlight}{
  on line,              
  colback=gray!20,     
  colframe=gray!20,      
  boxrule=0pt,
  arc=1mm,               
  left=0.5mm, right=0.5mm,
  top=0.2mm, bottom=0.2mm, 
  enhanced                 
}

\begin{document}
\pagestyle{plain}

\title{Are We in the AI-Generated Text World Already? \\ Quantifying and Monitoring AIGT on Social Media}
\date{}
\author{
Zhen Sun\textsuperscript{1}\thanks{Equal contribution.}  \ \ \ 
Zongmin Zhang\textsuperscript{1}\footnotemark[1]  \ \ \ 
Xinyue Shen\textsuperscript{2}  \ \ \ 
Ziyi Zhang\textsuperscript{1}  \ \ \ 
\\
\\
Yule Liu\textsuperscript{1}     \ \ \
Michael Backes\textsuperscript{2}  \ \ \ 
Yang Zhang\textsuperscript{2}  \ \ \ 
Xinlei He\textsuperscript{1}\thanks{Corresponding author (\href{mailto:xinleihe@hkust-gz.edu.cn}{xinleihe@hkust-gz.edu.cn}).} \ \ \
\\
\\
\textsuperscript{1}\textit{The Hong Kong University of Science and Technology (Guangzhou)} \ \ \ 
\\
\\
\textsuperscript{2}\textit{CISPA Helmholtz Center for Information Security} \ \ \ 
\\
\\
}

\maketitle

\begin{abstract}

Social media platforms are experiencing a growing presence of AI-Generated Texts (AIGTs).
However, the misuse of AIGTs could have profound implications for public opinion, such as spreading misinformation and manipulating narratives.
Despite its importance, it remains unclear how prevalent AIGTs are on social media. 
To address this gap, this paper aims to quantify and monitor the AIGTs on online social media platforms. 
We first collect a dataset (\Dataset) with around $2.4M$ posts from $3$ major social media platforms: Medium, Quora, and Reddit. 
Then, we construct a diverse dataset (\Bench) to train and evaluate AIGT detectors.
\Bench combines popular open-source datasets and our AIGT datasets generated from social media texts by $12$ LLMs, serving as a benchmark for evaluating mainstream detectors.
With this setup, we identify the best-performing detector (\Detector).
We then apply \Detector to \Dataset to track AIGTs across social media platforms from January 2022 to October 2024, using the AI Attribution Rate (AAR) as the metric.
Specifically, Medium and Quora exhibit marked increases in AAR, rising from $1.77\%$ to $37.03\%$ and $2.06\%$ to $38.95\%$, respectively.
In contrast, Reddit shows slower growth, with AAR increasing from $1.31\%$ to $2.45\%$ over the same period.
Our further analysis indicates that AIGTs on social media differ from human-written texts across several dimensions, including linguistic patterns, topic distributions, engagement levels, and the follower distribution of authors.
We envision our analysis and findings on AIGTs in social media can shed light on future research in this domain.
Our code and dataset are publicly available.\footnote{Code \& Dataset: \url{https://github.com/TrustAIRLab/AIGT_on_Social_Media}.}
\end{abstract}

\section{Introduction}

The rapid development of Large Language Models (LLMs) has markedly enhanced the quality of AIGTs, enabling the use of models like GPT-3.5~\cite{openai_chatgpt} in daily life to produce high-quality texts, such as in academic writing~\cite{gruda2024three}, question-answering~\cite{kamalloo2023evaluating}, and translation~\cite{wang2023document}. 
These AIGTs are often indistinguishable from Human-Written Texts (HWTs), presenting AIGT detection as a crucial yet challenging task for effective classification.
On social media platforms, the use of LLMs to answer questions can contribute to the spread of misinformation~\cite{zhou2023synthetic}. 
Furthermore, AIGTs may be deliberately used for information manipulation or the dissemination of fake news, potentially resulting in serious societal impacts~\cite{hanley2024machine}.
To better understand the prevalence of AIGTs on social media platforms, we aim to quantify and monitor their presence, addressing the question: \textbf{On social media, are we already interacting with AI-generated texts?}

\begin{figure}
    \centering
    \includegraphics[width=1\linewidth]{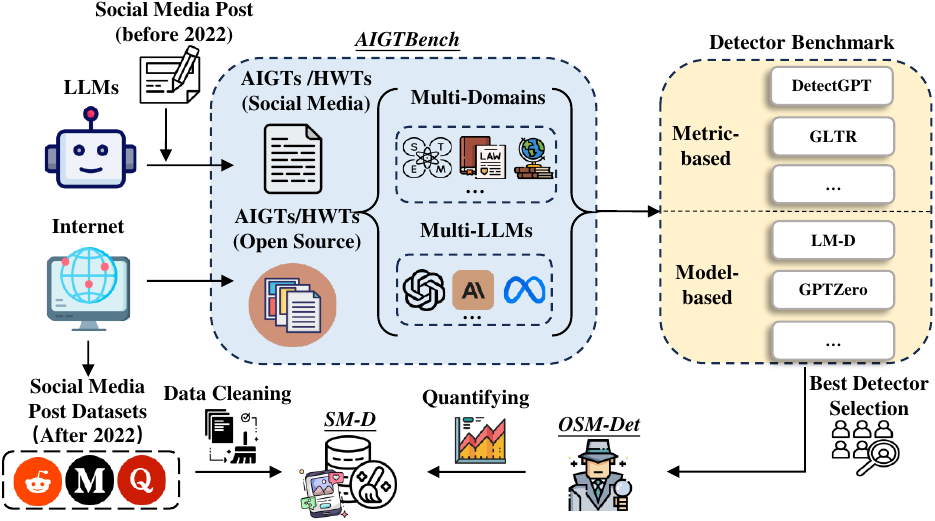}
    \caption{Pipeline for quantifying AIGTs on social media: \Dataset (2.4M posts), \Bench (training benchmark), \Detector (optimal detector).}
    \label{fig:pipeline}
\end{figure}

Currently, numerous detectors have been developed to detect AIGTs.
According to the MGTBench~\cite{he2023mgtbench}, these detectors are broadly divided into two categories: metric-based~\cite{gehrmann2019gltr, mitchell2023detectgpt} and model-based detectors~\cite{ippolito2019automatic,solaiman2019release,bhattacharjee2023conda}, some of which have shown high accuracy and robustness. 
While these detectors have been applied in controlled settings, recent studies have explored their effectiveness in real-world scenarios. 
\citet{hanley2024machine} conduct AIGT detection on news website articles, with a primary focus on content generated by GPT-3.5 and others from Turing benchmark, which includes various pre-2022 models~\cite{uchendu2021turingbench}.
Furthermore, \citet{liu2023detectability} detects ChatGPT-generated content on arXiv papers.
However, academic and news writing are formal and tailored to specific audiences, whereas social media content is more interactive, making it a better domain for observing AIGTs' impact on daily life. 
Moreover, previous studies do not account for recent popular LLMs, while we consider a broader range of models in our efforts to detect AIGTs on social media.

To quantify and monitor AIGTs on social media, we collect textual data from $3$ popular platforms spanning January 1, 2022, to October 31, 2024, as most LLMs are released after 2022. 
After data preprocessing, we obtain $1,170,821$ posts from Medium, $245,131$ answers from Quora, and $982,440$ comments from Reddit.
We name it as \Dataset, short for \underline{S}ocial \underline{M}edia \underline{D}ataset.

To identify the most effective detector, we construct a dataset named \Bench, which consists of public AIGT/Supervised-Finetuning (SFT) datasets and our own AIGT datasets generated from social media data.
\Bench includes AIGTs generated by $12$ different LLMs, such as GPT Series~\cite{openai_gpt4o_mini}) and Llama Series~\cite{touvron2023llama,touvron2023llama2,dubey2024llama}), totaling around $28.77M$ AIGT and $13.55M$ HWT samples. 
We then benchmark AIGT detectors on \Bench and leverage the best-performing detector as our primary detector, which achieves an accuracy of $0.979$ and an F1-score of $0.980$.
To better reflect its application in detecting AIGTs on online social media, we rename it as \Detector (\underline{O}nline \underline{S}ocial \underline{M}edia \underline{Det}ector).

Based on \Detector, we quantify and monitor the texts across the $3$ platforms and use the AI Attribution Rate (AAR) to represent the rate of posts classified as AI-generated (The pipeline is shown in~\Cref{fig:pipeline}).
We observe several noteworthy phenomena:
(1) \textbf{A sharp rise in AI-generated content begins in December 2022, with distinct AAR trends emerging across platforms.}
Before December 2022, the AAR across platforms remains stable.
However, starting in December, Medium and Quora show significant surges, while Reddit shows only a slight increase.
This suggests the widespread and diverse LLM adoption on social media;
(2) \textbf{Linguistic analysis shows similar AAR trends and exhibits stylistic features in AIGTs/HWTs.}
Based on the word-level analysis, we find that the usage trend of top-frequency AI-preferred words aligns closely with LLM adoption trends.
With sentence-level analysis, we also reveal that AIGTs tend to be more objective and standardized, whereas HWTs are more flexible and informal;
(3) \textbf{Technology-related topics drive higher AARs on Medium.}
Topics like ``Technology'' and ``Software Development'' show the highest AARs, indicating that users with a strong technical background are more likely to adopt LLMs;
(4) \textbf{Predicted HWTs receive more engagement than AIGTs.}
On Medium, the content predicted as HWTs receives more average ``Likes'' and ``Comments'' than AIGTs.
This suggests that users are more inclined to engage with HWTs;
and (5) \textbf{Authors with fewer followers are more likely to produce AIGTs.}
On Medium, users with no more than one thousand followers tend to produce content that has the highest mean AAR at $54.02\%$.
In contrast, as the follower count increases, the AAR gradually shifts toward the lower range ($\leq25.00\%$).

Our contributions are summarized as follows:

\begin{itemize}
    \item We are the first to conduct a systematic study to quantify, monitor, and analyze AIGTs on social media.
    To achieve this, we collect a large-scale dataset \Dataset, which includes around $2.4M$ posts from three platforms, spanning from January 2022 to October 2024.
    \item We construct \Bench, a dataset for benchmarking AIGT detectors.
    \Bench can be divided into two parts: one derived from open-source datasets and the other generated by $12$ LLMs based on platform-specific characteristics.
    Leveraging \Bench, we identify the most effective AIGT detector, \Detector.
    \item  Our research reveals a remarkable increase in AAR on social media after the widespread adoption of LLMs. Moreover, this trend varies markedly across different platforms.
    \item We conduct an in-depth analysis of the characteristics of AIGTs and HWTs through \emph{linguistic analysis} and \emph{multidimensional analysis of posts}, revealing differences in lexical patterns, topic distributions, engagement levels, and the follower distributions of authors. These analyses provide valuable insights for future research.
\end{itemize}

\section{Related Work}

The growth in model parameters and training data has recently empowered LLMs to demonstrate exceptional language processing capabilities~\cite{zhao2023survey}. 
Since then, LLMs have gradually gained popularity, like GPT-4~\cite{openai2023gpt4} and Llama~\cite{touvron2023llama}, enabling users to generate high-quality texts effortlessly.
Yet, LLMs exhibit multiple inherent vulnerabilities~\cite{He2025AISecuritySurvey,sun2024peftguard,liu2024quantized,zheng2025cl} and have raised concerns about potential misuse, such as fake news generation~\cite{zellers2019defending}, academic misconduct~\cite{vasilatos2023howkgpt}, hate speech generation~\cite{SWQBZZ25}, and performance degradation of training LLMs using AI content~\cite{briesch2023large}, making the detection of AIGTs (also known as machine-generated texts) increasingly important~\cite{fraser2024detecting}.
\citet{he2023mgtbench} introduce MGTBench for standardizing the evaluation of different LLMs and experimental setups within the AIGT detectors. 
They broadly categorize the detectors into two main types: metric-based and model-based detectors. 
Metric-based detectors use pre-defined metrics, such as log-likelihood, to capture the characteristics of texts~\cite{gehrmann2019gltr, mitchell2023detectgpt, su2023detectllm}.
In contrast, model-based detectors rely on trained models to distinguish between AIGTs and HWTs~\cite{solaiman2019release, guo2023close, bhattacharjee2023conda, liu2023detectability, ippolito2019automatic, li2024mage}. 
More introduction refer to~\Cref{sec:detector_intro}.

Besides, some researchers have applied detectors to text detection in real-world scenarios.
\citet{hanley2024machine} train a detector using data generated by the ChatGPT and Turing benchmark model and conduct tests on multiple news websites. 
Their study reveals that, from January 1, 2022, to May 1, 2023, the proportion of synthetic articles increased on news sites.
\citet{liu2023detectability} also conduct detection on arXiv and find a significant rise in the proportion of papers using ChatGPT-generated content, reaching $26.1\%$ by December 2023.
In contrast to their detection targets, we focus on detecting AIGTs on social media platforms and covering a broader range of LLMs.
\citet{macko2024multisocial} construct a multilingual dataset based on instant messaging and social interaction platforms such as Telegram and Discord, using it to compare the performance of existing detectors. 
In contrast, our research focuses on providing an in-depth temporal analysis of AIGTs on content-driven social platforms like Medium, Quora, and Reddit. 

\section{Data Collection}
\label{sec:data_collection}

In this section, we elaborate on the data collection process, which primarily includes two datasets: the social media dataset (\Dataset) and the detector training dataset (\Bench).

\subsection{\Dataset (Social Media Dataset)}
\label{sec:social_media_dataset}

\begin{table}[h]
\centering
\resizebox{0.48\textwidth}{!}{
\begin{tabular}{@{}lrrc@{}}
\toprule
Dataset & \# Posts & \# Filtered Posts & Time Range            \\ \midrule
Medium  & $1,416,208$       & $1,170,821$               & January 1, 2022-October 31, 2024 \\
Quora   & $445,864$       & $245,131$               & January 1, 2022-October 31, 2024 \\
Reddit  & $1,019,261$       & $982,440$               & January 1, 2022-July 31, 2024 \\ \bottomrule
\end{tabular}}
\caption{Overview of the Medium, Quora, and Reddit datasets.}
\label{data_collection}
\end{table}

Unlike previous research, we focus on social media platforms, including Medium, Quora, and Reddit, emphasizing content creation, sharing, and discussion. The introduction of platforms is in~\Cref{sec:social_media_platoforms}.
These platforms stand out for hosting longer, more detailed posts where users emphasize the depth and quality of the information they share.
As shown in~\Cref{data_collection}, we collect data from these social media platforms from January 1, 2022 to October 31, 2024.
We consider this part as our social media dataset for analysis.

For each platform, the detection targets are determined based on their distinct characteristics.
On Medium, a blog hosting platform, we extract both the titles and contents of articles, treating the entire article as the detection target.
On Quora, a question-and-answer platform, we select the corresponding answers to questions as the detection target.
Similarly, on Reddit, which is known for its user-driven discussions, we also choose the response content as the detection target.
Furthermore, we apply data filtering with the rules described in~\Cref{sec:data_preprocessing}.

\subsection{\Bench (Detector Training Dataset)}
\label{sec:training_dataset}

\begin{figure}[h!]
    \centering
    \includegraphics[width=0.8\linewidth]{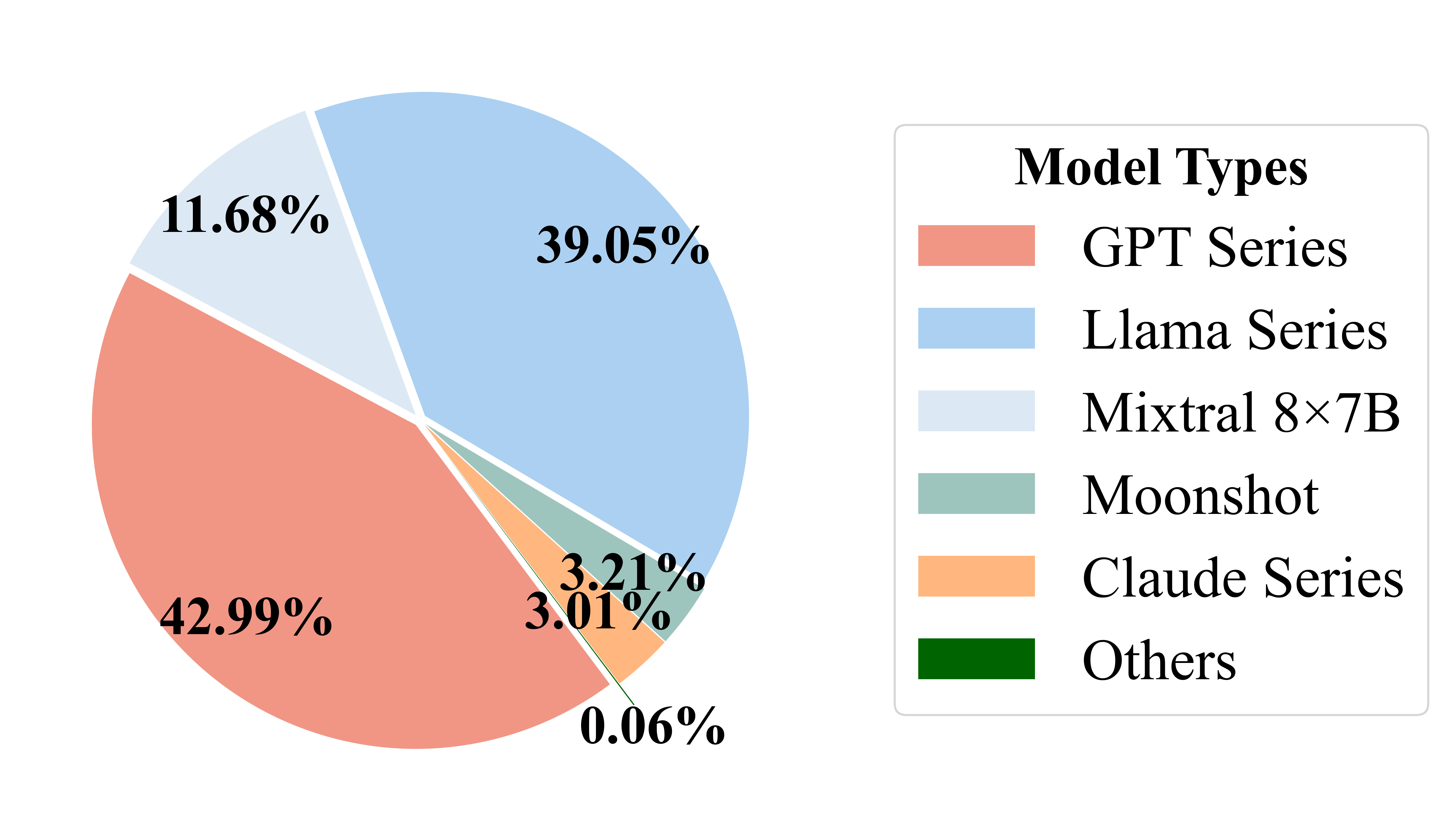}
    \caption{Proportion of total sentences various LLMs, with ``Others'' including Alpaca 7B and Vicuna 13B.}
    \label{fig:Proportion-token}
\end{figure}

To train the AIGT detectors, we consider two parts of the data.
First, we consider $6$ publicly available AIGT datasets and $5$ common SFT datasets to form the training dataset (see~\Cref{tab:datasets_tokens_1,tab:datasets_tokens_2} for dataset statistics and~\Cref{sec:training_detector_intro} for more details).
Second, to increase the detector's generalization capabilities on social media, we additionally collect data from the $3$ social media platforms ranging from January 1, 2018, to December 31, 2021, as verified by an ablation study demonstrating that these new subsets fill a gap that older benchmarks missed (see \Cref{sec:ablation_study}). 
We classify this data as HWTs, given that most LLMs had not been published during this period.
We also design different LLMs writing tasks to generate AIGTs that align with the characteristics of platforms (\Cref{tab:generated} describes the statistics details).

For Medium, which is primarily used for sharing articles and blogs, the core tasks are centered on writing.
We design two LLM writing tasks: (1) polish articles to create polished versions; (2) based on the article's title and summary, directing the LLM to generate complete article content, thereby simulating a writing scenario.
For Quora and Reddit, which mainly focus on question answering and user interaction, we design two tasks: (1) polish texts like Medium and (2) query LLM directly answer questions, simulating a user interaction scenario.
Detailed prompts are provided in~\Cref{sec:task_prompt}.

Overall, the datasets used for training our detector and the distribution of LLM series are shown in~\Cref{fig:Proportion-token}.
This dataset includes $12$ different LLMs, with a detailed introduction provided in~\Cref{sec:model_intro}.
Within these datasets, the two most prevalent model series are the GPT Series, which accounts for $42.99\%$, and the Llama series, which represents $39.05\%$. 
GPT Series is the most widely used proprietary model and has played a pivotal role in the evolution of generative AI.
As of January 2023, approximately $13M$ users interact daily with GPT-3.5~\cite{10221755}. 
The Llama series models also have significant influences, as the report indicates that downloads of Llama models on the Hugging Face platform have nearly reached around $350M$~\cite{meta2024llamareport}. 
Therefore, these two model series are the primary focus of our dataset.
During the data generation process, we notice that certain samples contain textual noise, like irrelevant or redundant information.
To maintain data quality, we implement some data processing strategies (see~\Cref{sec:data_preprocessing} for details).

\begin{table*}[h!]
    \centering
    \resizebox{\textwidth}{!}{
    \begin{tabular}{l|cccccccc|cccccc}
        \toprule
         & \multicolumn{8}{c|}{Metric-based} & \multicolumn{6}{c}{Model-based} \\ 
        \cmidrule(lr){2-9} \cmidrule(lr){10-15}
                                & \makecell{Log-\\Likelihood} & Rank & \makecell{Log-\\Rank} & Entropy & GLTR & LRR & DetectGPT & NPR 
                                & \makecell{OpenAI \\ Detector} & \makecell{ChatGPT \\ Detector} & ConDA & GPTZero & CheckGPT & LM-D \\ 
        \midrule
        Accuracy               & $0.730$          & $0.618$ & $0.713$    & $0.650$   & $0.704$ & $0.680$ & $0.686$     & $0.658$
                                & $0.615$           & $0.686$            & $0.972$ & $0.933$  & $0.966$    & \bm{$0.979$} \\ 
        F1-score               & $0.754$          & $0.730$ & $0.741$    & $0.697$   & $0.733$ & $0.660$ & $0.659$     & $0.639$
                                & $0.484$           & $0.602$            & $0.973$ & $0.930$  & $0.966$    & \bm{$0.980$} \\ 
        \bottomrule
    \end{tabular}
    }
    \caption{Performance of detectors on \Bench. The F1-score corresponds to the AI class.}
    \label{tab:benchmark_performance}
\end{table*}

\section{Experimental Settings}

\subsection{Datasets}

As mentioned in~\Cref{sec:data_collection}, we collect the social media dataset (\Dataset) and the detector training dataset (\Bench).
\Dataset refers to the social media dataset that we conduct the quantification, with more details provided in~\Cref{sec:social_media_dataset}.
\Bench is the benchmark for AIGT detectors, which includes AIGTs generated by $12$ different LLMs, as described in~\Cref{sec:training_dataset}.
We randomly divide \Bench into training, validation, and test sets in a $7:1:2$ ratio.
Specifically, the distribution of token lengths in the training, validation, and test set are shown in~\Cref{fig:train_token_length_distribution}. 

\subsection{AIGT Detectors}

Following the experimental setup of MGTBench~\cite{he2023mgtbench}, we evaluate 14 detectors.
For metric-based detectors, we consider LogLikelihood, Rank, LogRank, Entropy, GLTR, LRR, DetectGPT, and NPR~\cite{solaiman2019release, gehrmann2019gltr, mitchell2023detectgpt}.
We choose the GPT-2 medium~\cite{radford2019language} as the base model, given its good detection performance at limited computational costs.

During the detection process, we initially use the GPT-2 medium to extract multiple metrics, including log-likelihood and log-rank.
Based on these extracted metrics, we train logistic regression models to enhance the accuracy of predictions.
For the model-based detectors, we consider both pre-trained detectors and fine-tuned models with the \Bench, that is, OpenAI Detector~\cite{solaiman2019release}, ChatGPT Detector~\cite{guo2023close}, ConDA~\cite{bhattacharjee2023conda}, GPTZero~\cite{gptzero}, CheckGPT~\cite{liu2023detectability}, and LM-D~\cite{ippolito2019automatic}.
Specifically, for the OpenAI Detector and ChatGPT Detector, we consider their pre-trained version and select the RoBERTa-base model as it demonstrates stable performance across multiple detection tasks and typically provides better detection results. 
For ConDA and LM-D, we choose the Longformer-base-4096 model as the base model and fine-tune it with the \Bench. All of them have a learning rate of 1e-5, a batch size of 16, and the AdamW optimizer.
For GPTZero, we directly use its commercial API.
For CheckGPT, we retrain the original training framework~\cite{liu2023detectability}.

\subsection{Evaluation Metrics}

We use accuracy and F1-score as the evaluation metrics to evaluate the performance of different detectors, which are common standards in AIGT detection tasks. 
Besides, we introduce two metrics \textbf{AI Attribution Rate} (AAR) and \textbf{False Positive Rate} (FPR) for quantification analysis.
The AAR indicates the proportion of texts that the model predicts as AI-generated, while the FPR denotes the proportion of HWTs misclassified as AIGTs.

To assess word usage, we compute the \textbf{normalized term frequency} (NTF) as:

\begin{equation}
\text{NTF}(t,d) = \frac{f_{t,d}}{N \cdot \sum_{t' \in d} f_{t',d}},
\end{equation}
where $f_{t,d}$ is the frequency of word $t$ in document $d$, $\sum_{t' \in d} f_{t',d}$ accounts for all words in $d$, and $N$ is the total occurrences of $t$ across all documents.

\section{Evaluation}

\subsection{Benchmarking Detectors}

This section compares different AIGT detectors on the test set of the \Bench.
Illustrated in~\Cref{tab:benchmark_performance}, the metric-based detectors perform poorly. 
The F1-scores for Log-Likelihood, Rank, Log-Rank, and Entropy are $0.754$, $0.730$, $0.741$, and $0.697$, respectively. 
These low scores indicate that metric-based detectors face limitations in handling complex, multi-source datasets and struggle to capture subtle textual features effectively.

Regarding model-based detectors, we observe that both OpenAI Detector and ChatGPT Detector perform worse than some metric-based detectors. 
Specifically, OpenAI Detector has an F1-score of only $0.484$, with relatively low accuracy. 
This underperformance may be due to the detector being fine-tuned using GPT-2 output, which struggles to adapt to more complex data generated by modern LLMs, such as the Llama and Claude Series.

Notably, LM-D and ConDA outperform the others.
ConDA achieves an accuracy of $0.972$, while the LM-D performs even better, with an accuracy of $0.979$ and an F1-score of $0.980$, making it the most effective detector. 
Based on these benchmark results, we consider LM-D as the most effective detection method and name LM-D fine-tuned on \Bench as \Detector, which is subsequently used to quantify and monitor the AAR in social media dataset (\Dataset).
More details on performance across different platforms and all text lengths in \Dataset are shown in~\Cref{sec:detailed_performance}.

\begin{table}[h!]
\centering
\footnotesize
\begin{tabular}{@{}ccc@{}}
\toprule
\textbf{Platform}   & \textbf{\# text (Human)} & \textbf{FPR}    \\
\midrule 
Medium     & $116,303$ & $1.82\%$ \\ 
Quora      & $101,145$ & $1.36\%$ \\
Reddit     & $53,321$ & $1.70\%$ \\
\bottomrule 
\end{tabular}
\caption{FPR of \Detector on social media platforms.}
\label{tab:text_analysis}   
\end{table}

\subsection{Generalizability of \Detector}
\label{sec:Generalizability}

\Bench is a comprehensive dataset that contains multi-source, multi-domain, and multi-LLM data. 
The diversity of this dataset enhances the generalizability of detectors in real-world (in-the-wild) environments. 
\Detector, on the other hand, is the optimal detection model trained on \Bench.

In this section, we evaluate the generalizability of \Detector from three perspectives: AIGTs produced with different generation parameters, AIGTs of social media generated by unseen models, and tests in the wild.

\begin{figure}[h!]
    \centering
    \includegraphics[width=0.40\textwidth]{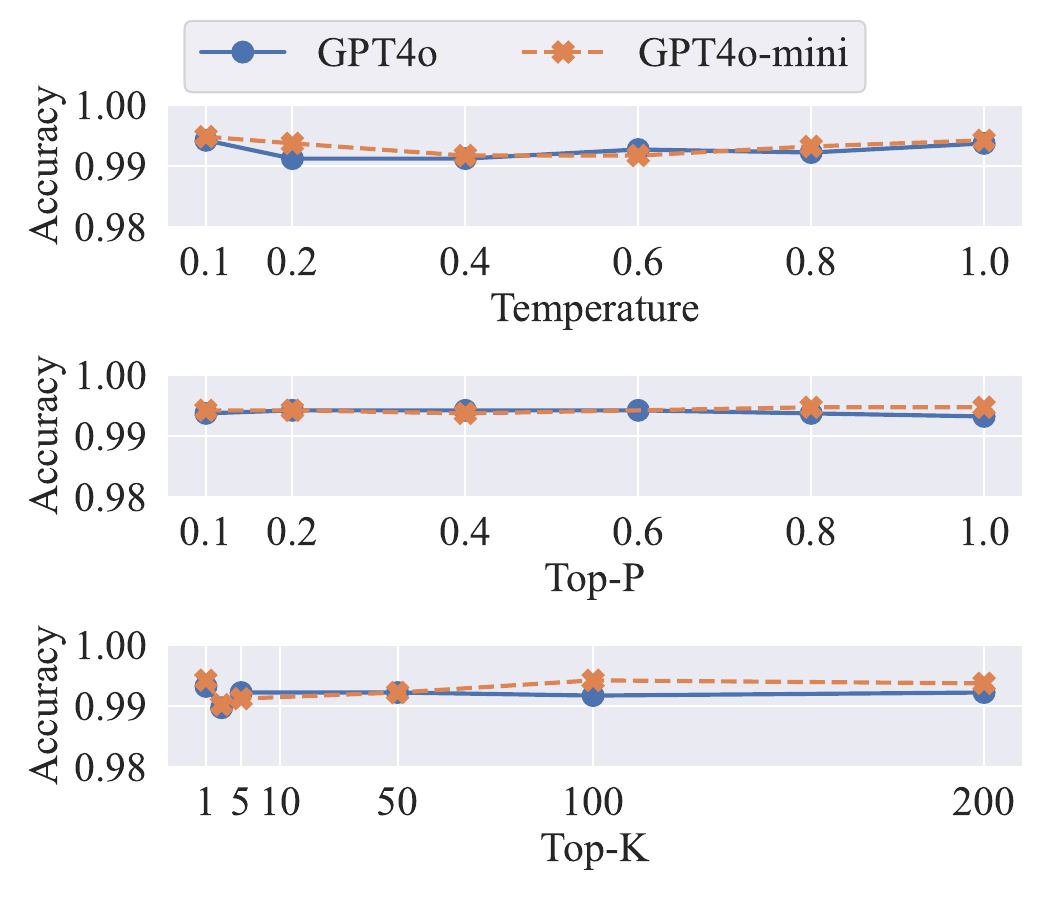}
    \caption{Impact of different generation parameters on AIGT detection accuracy.}
    \label{fig:aigt_accuracy}
\end{figure}

\mypara{Different Generation Parameters}
To investigate whether \Detector can effectively detect AIGTs generated with different generation parameters, we randomly sample $5,000$ HWTs from the \Bench and apply the same prompt to refine them using different generation parameters (including temperature, top-p, and top-k). 
The models used for this experiment are GPT4o and GPT4o-mini.

As shown in \Cref{fig:aigt_accuracy}, \Detector maintains an accuracy of over $0.99$ across the entire range of temperature settings (0.1 to 1.0). 
Top-P (0.1 to 1.0) and Top-K (1 to 200) show a similar trend. 
This indicates that \Detector demonstrates strong generalizability when detecting AIGTs generated with different parameters.

\begin{table}[ht!]
    \centering
    \scalebox{0.9}{
    \footnotesize
    \begin{tabular}{@{}ccc@{}}
    \toprule
    \textbf{Model} & \textbf{Accuracy} & \textbf{F1-score} \\ 
    \midrule
    Deepseek-V3              & 0.986 & 0.993 \\
    GLM-4-Flash              & 0.997 & 0.998 \\
    Gemini-1.5-Flash         & 0.938 & 0.952 \\
    Gemini-2.0-Flash         & 0.984 & 0.992 \\
    Yi-1.5-34B               & 0.999 & 0.999 \\
    InternVL2.5-8B           & 0.925 & 0.958 \\
    Dolphin3.0-Llama3.1-8B   & 0.996 & 0.998 \\
    Llama3-OpenBioLLM-8B     & 0.960 & 0.980 \\
    Xwin-LM-13B-V0.2         & 0.996 & 0.998 \\
    \bottomrule
    \end{tabular}}
    \caption{Performance of \Detector on AIGTs generated from unseen LLMs based on social media data from \Bench.}
    \label{tab:unseen_llm_on_social_media}
\end{table}

\mypara{AIGTs of Social Media Generated By Unseen Models}
To investigate the generalizability of \Detector on social media AIGTs generated by unseen models, we selected $6$ pre-trained models, including Deepseek-V3~\cite{liu2024deepseek}, GLM-4-Flash~\cite{GLM4API}, Gemini-1.5-Flash~\cite{GeminiFlash}, Gemini-2.0-Flash~\cite{GeminiFlash}, Yi-1.5-34B~\cite{young2024yi}, and InternVL2.5-8B~\cite{InternVL2_5-8B}.
Additionally, we include three fine-tuned models based on the LLaMA series: Dolphin3.0-Llama3.1-8B~\cite{Dolphin3.0-Llama3.1-8B}, Llama3-OpenBioLLM-8B~\cite{Llama3-OpenBioLLM-8B}, and Xwin-LM-13B-V0.2~\cite{Xwin-LM-13B-V0.2}. 
Since none of these models were included in \Bench, they are considered unseen to \Detector. 
We also apply the same polishing process to the previously selected $5,000$ data samples.

From~\Cref{tab:unseen_llm_on_social_media}, we observe that \Detector maintains strong detection performance across these unseen models. 
The lowest performance was recorded for InternVL2.5-8B, yet it still achieve an accuracy of $0.925$ and an F1-score of $0.958$.
This demonstrates that \Detector exhibits strong generalization capability when detecting AIGTs generated by previously unseen LLMs.

\mypara{Test In the Wild}
To test \Detector in the wild, we randomly select datasets from the huggingface platform for evaluation. 
These datasets are in two main categories: unseen models and unseen domains, neither of which are included in \Bench.

As shown in~\Cref{tab:test_detector}, for the unseen model scenario, the test results align with previous findings, where \Detector maintains high accuracy and F1-score.
Similarly, in the unseen domain scenario, \Detector also demonstrates strong generalizability, achieving a minimum accuracy of $0.943$. 
This is consistent with the findings of~\citet{liu2024generalizationabilitymachinegeneratedtext}, which suggest that AIGT detectors exhibit generalizability across different domains.

\subsection{Evaluation on Social Media Platforms}
\label{sec:eval_smp}

As shown in~\Cref{tab:text_analysis}, \Detector achieves False Positive Rates (FPR) of $1.82\%$, $1.36\%$, and $1.70\%$ on Medium, Quora, and Reddit, respectively, while achieving a benchmark F1-score of $0.980$ (see~\Cref{tab:benchmark_performance}).
These results highlight \Detector's low misclassification rate and high overall accuracy, making it a reliable choice for quantifying and monitoring AIGTs on social media.

\begin{figure}[h!]
    \centering
    \begin{subfigure}[t]{\linewidth}
        \centering
        \includegraphics[width=0.92\linewidth]{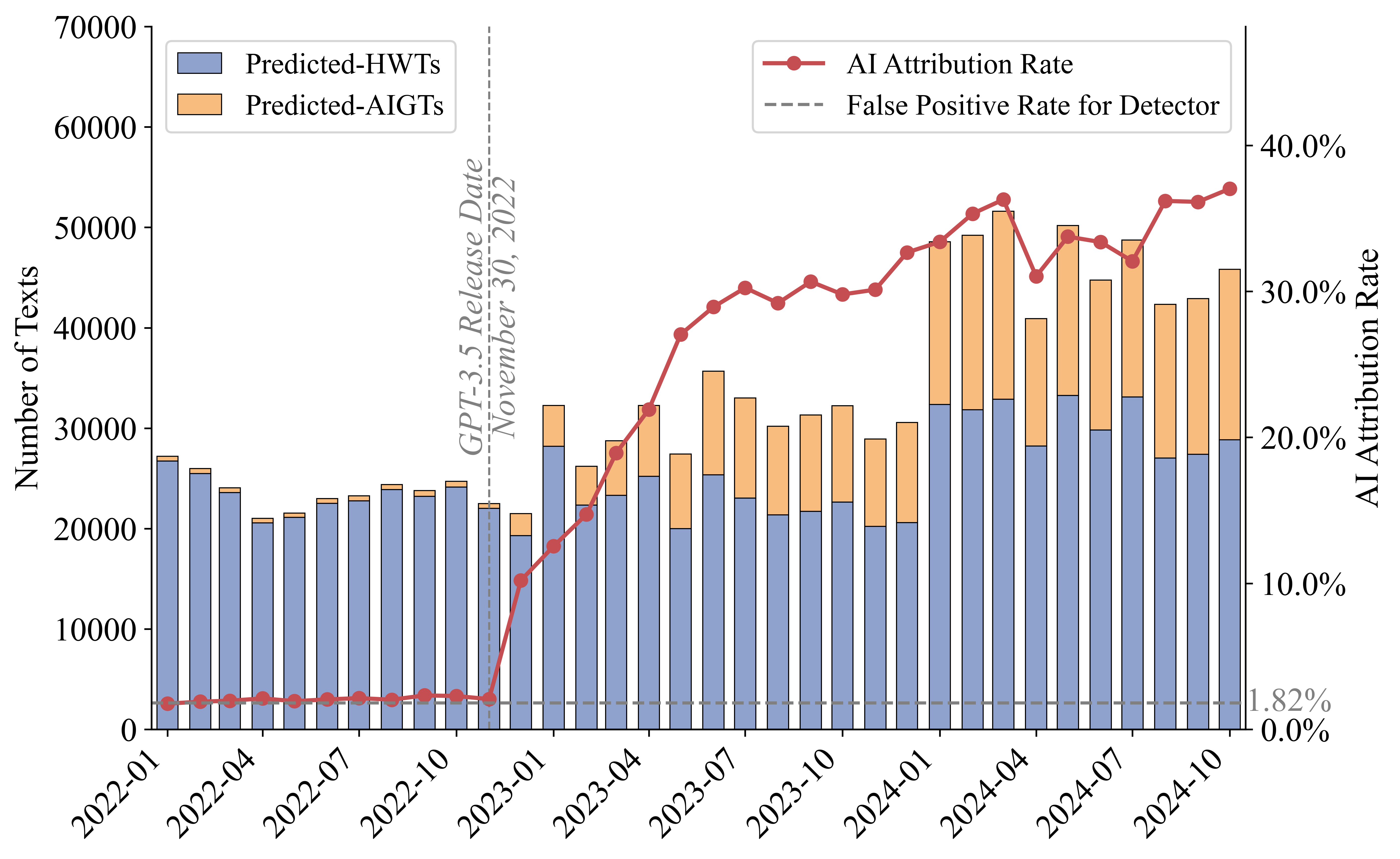}
        \caption{AAR Trends on Medium from January 1, 2022, to October 31, 2024.}
        \label{fig:medium_overall_combined}
    \end{subfigure}

    \begin{subfigure}[t]{\linewidth}
        \centering
        \includegraphics[width=0.92\linewidth]{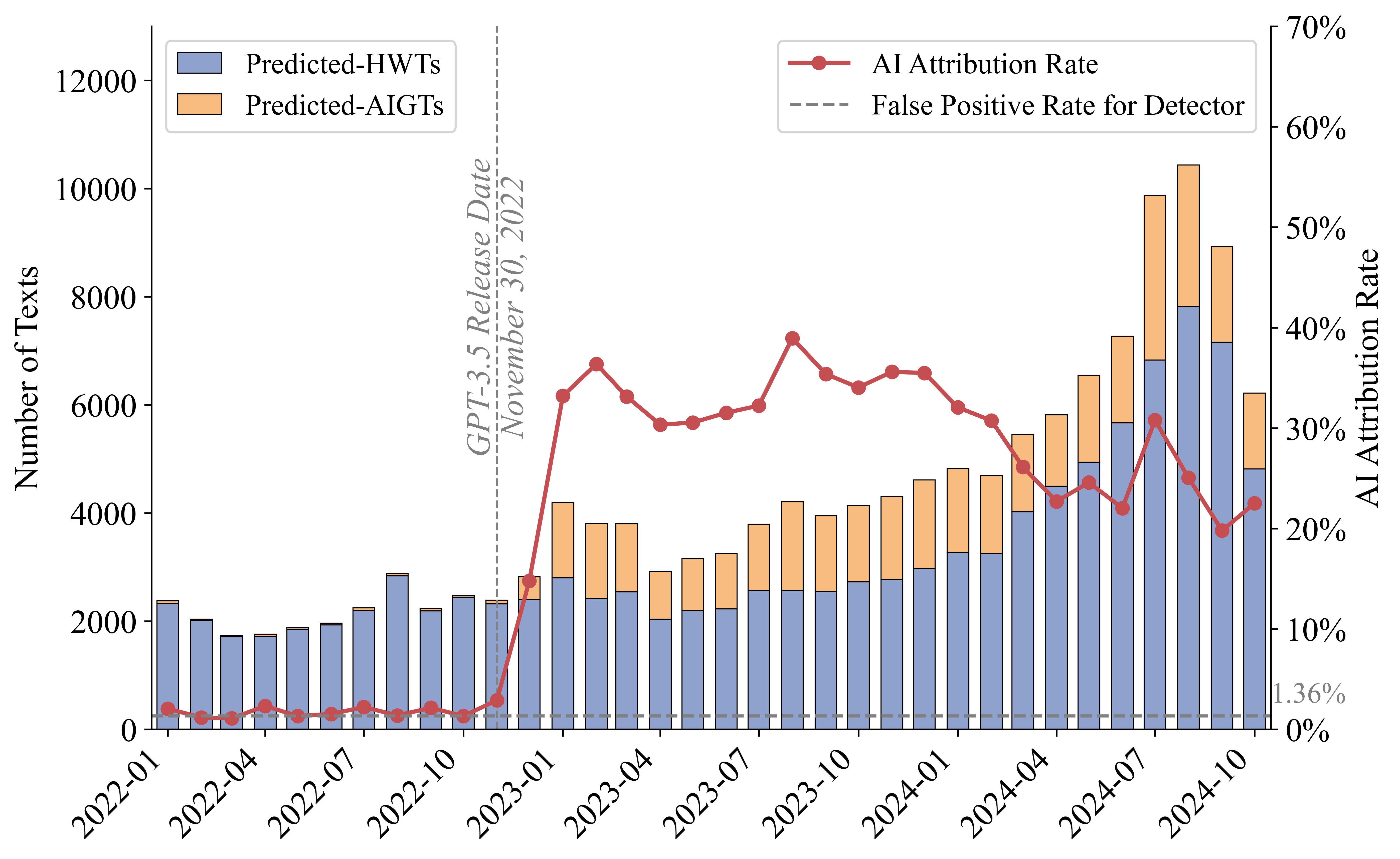}
        \caption{AAR Trends on Quora from January 1, 2022, to October 31, 2024.}
        \label{fig:quora_overall_combined}
    \end{subfigure}

    \begin{subfigure}[t]{\linewidth}
        \centering
        \includegraphics[width=0.92\linewidth]{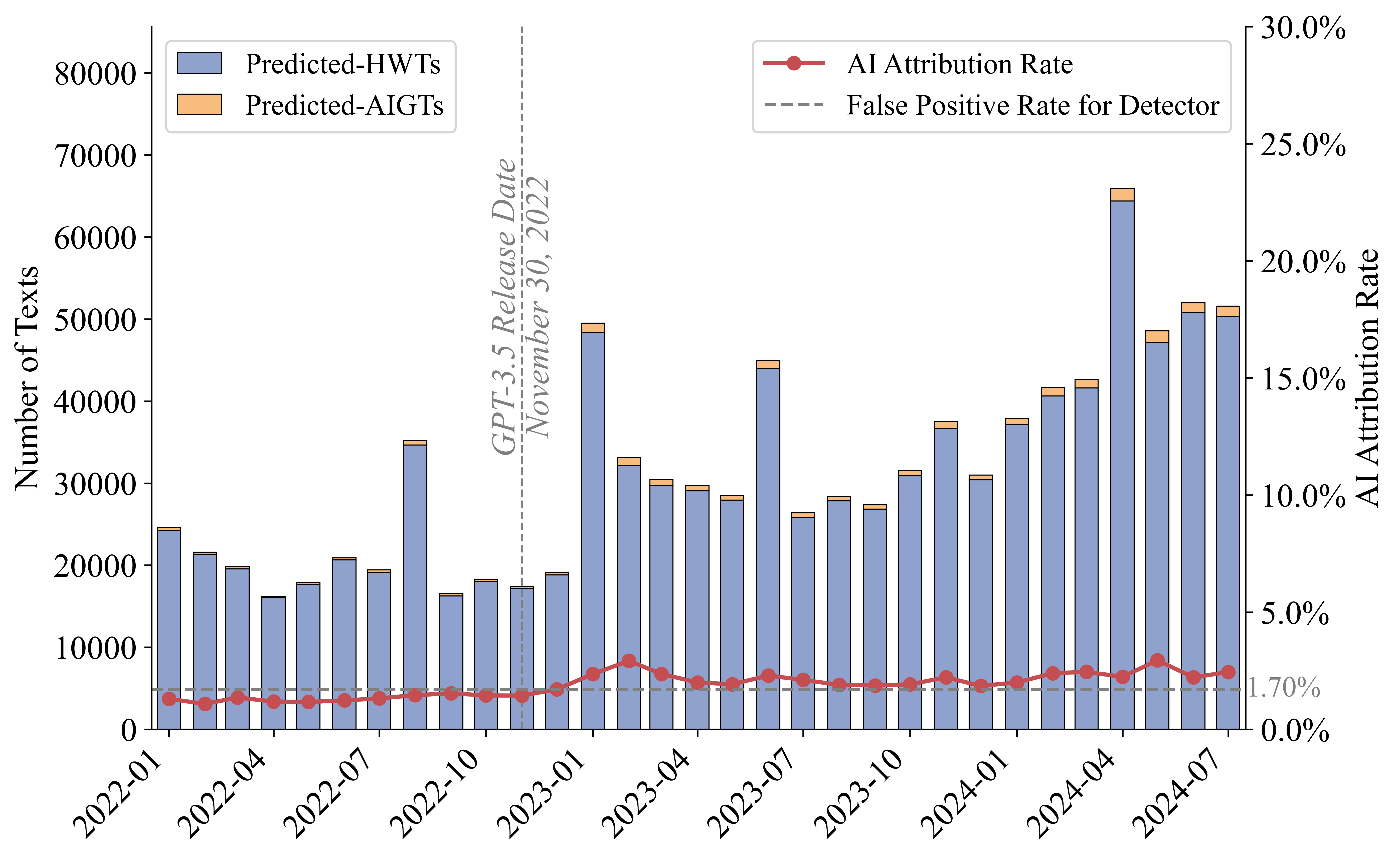}
        \caption{AAR Trends on Reddit from January 1, 2022, to July 31, 2024.}
        \label{fig:reddit_overall_combined}
    \end{subfigure}

    \caption{Comparison of AAR and FPR across Medium, Quora, and Reddit over different time periods.}
    \label{fig:overall_combined_comparison}
\end{figure}

\mypara{Evaluation on Medium}
\Cref{fig:medium_overall_combined} illustrates the trend of AAR on Medium from January 2022 to October 2024. 
From January 2022 to November 2022, the AAR remains stable, fluctuating around $1.82\%$. 
This suggests that, before the widespread adoption of GPT-3.5, creators mainly rely on original content with minimal dependency on LLM-generated content.
However, starting in December 2022, coinciding with the launch of GPT-3.5, the AAR begin to rise rapidly. 
Between December 2022 and July 2023, the AAR surges from $10.20\%$ to $30.24\%$, reflecting how the popularization of LLM technology significantly lowers the barriers of content generation, prompting Medium's creator community to widely adopt LLM-assisted content creation.
From August 2023 to July 2024, the AAR experiences slower growth, ranging between $29.20\%$ and $36.29\%$, with fluctuations stabilizing between $30.12\%$ and $33.75\%$.
This indicates that AIGTs have gradually become an integral part of the platform's creative ecosystem, serving as a critical component of content production.
From August 2024 to October 2024, the AAR further increased to $37.03\%$, reaching a new peak. 
This likely reflects the growing acceptance and reliance on LLM-assisted creation among content creators to enhance writing efficiency and quality.

Overall, from December 2022 to October 2024, the AAR on Medium has shown a continuous upward trend, underscoring the significant impact of LLM technology on content creation.

\mypara{Evaluation on Quora}
\Cref{fig:quora_overall_combined} displays the trend of AAR on Quora.
We observe that from January 2022 to October 2022, the AAR fluctuates but remains relatively low. 
After the release of GPT-3.5 in November 2022, the AAR slightly increases to $2.87\%$. 
Subsequently, starting in December 2022, the AAR markedly rises to $15.12\%$ and shows a clear upward trend in AIGTs, reaching a peak of $38.95\%$ in August 2023. 
From September 2023 to the first half of 2024, although the AAR remains high, it declines from the peak in early 2023 and gradually stabilizes between $22.03\%-30.79\%$ throughout 2024. 
This indicates that the behavior of Quora users in generating AI content is becoming more stable.
From June 2024, the AAR gradually decreases and reaches a low near $19.79\%$ between September and October 2024. 
The increase in AAR may be attributed to Quora's launch of its LLM platform, Poe, in 2023~\cite{quora_blog_poe}, which initially led to a rise in AI-generated content. 
However, as many Quora users found Poe's capabilities insufficient to meet their daily needs, the AAR likely declined following this initial surge, eventually stabilizing.

\mypara{Evaluation on Reddit}
\Cref{fig:reddit_overall_combined} shows the quantification analysis on Reddit from January 2022 to July 2024.
From January to November 2022, we observe that the AAR remains below the FPR, fluctuating around $1.30\%$, indicating that there is almost no AI-generated content on Reddit during this period. 
Following the release of GPT-3.5, the AAR begins to rise slightly, reaching $2.36\%$ in January 2023 and further increases to $2.93\%$ in February 2023. 
From March 2023 to July 2024, the AAR stabilizes at a low level, within the range of $1.86\%-2.95\%$.

Briefly, similar to Medium and Quora, AAR on Reddit shows an upward trend following the release of GPT-3.5, but it consistently maintains a lower level, indicating a lower dependency on LLMs among Reddit users.

\begin{figure}[h!]
    \centering
    \begin{subfigure}[t]{\linewidth}
        \centering
        \includegraphics[width=0.90\linewidth]{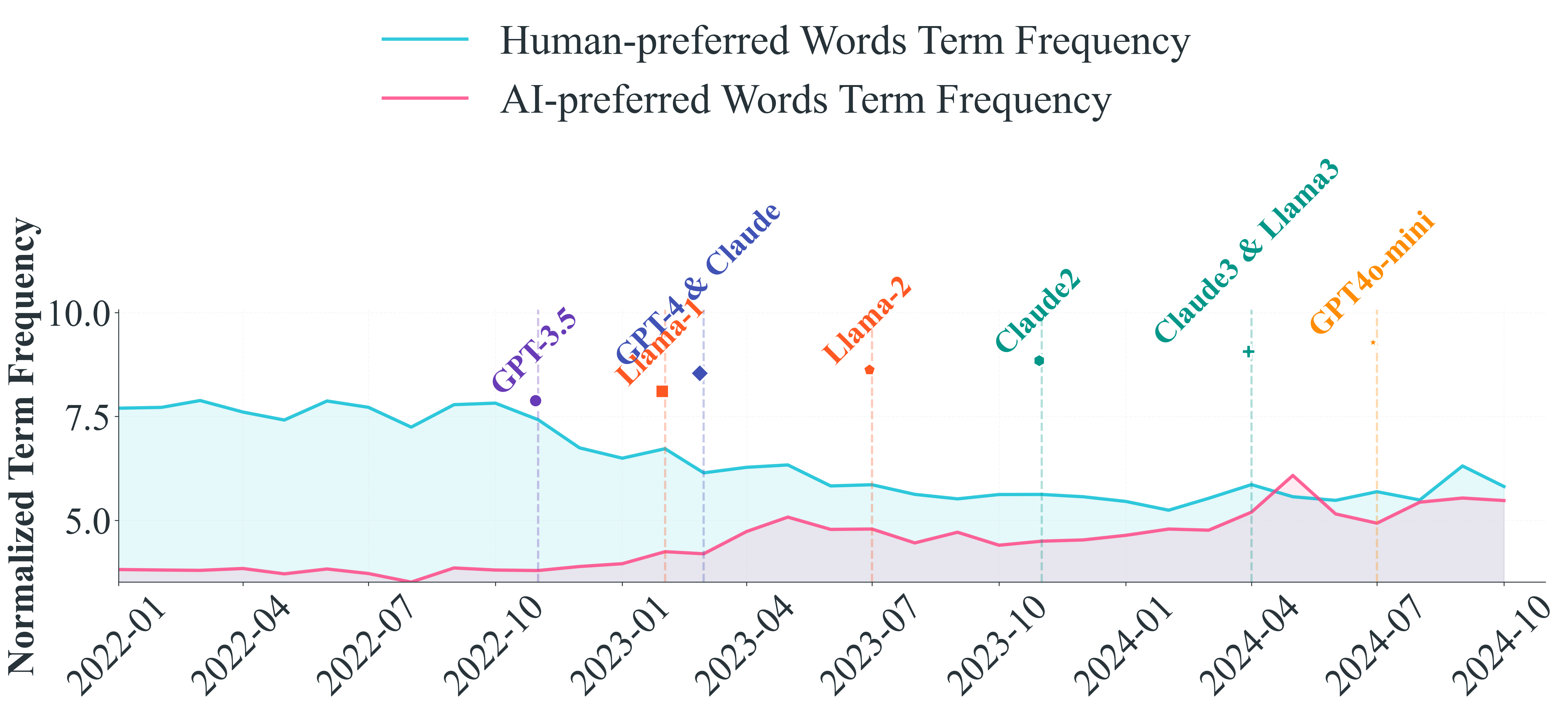}
        \caption{Word frequency trends on Medium from January 1, 2022, to October 31, 2024.}
        \label{fig:medium-words-trends-chart}
    \end{subfigure}
    \begin{subfigure}[t]{\linewidth}
        \centering
        \includegraphics[width=0.90\linewidth]{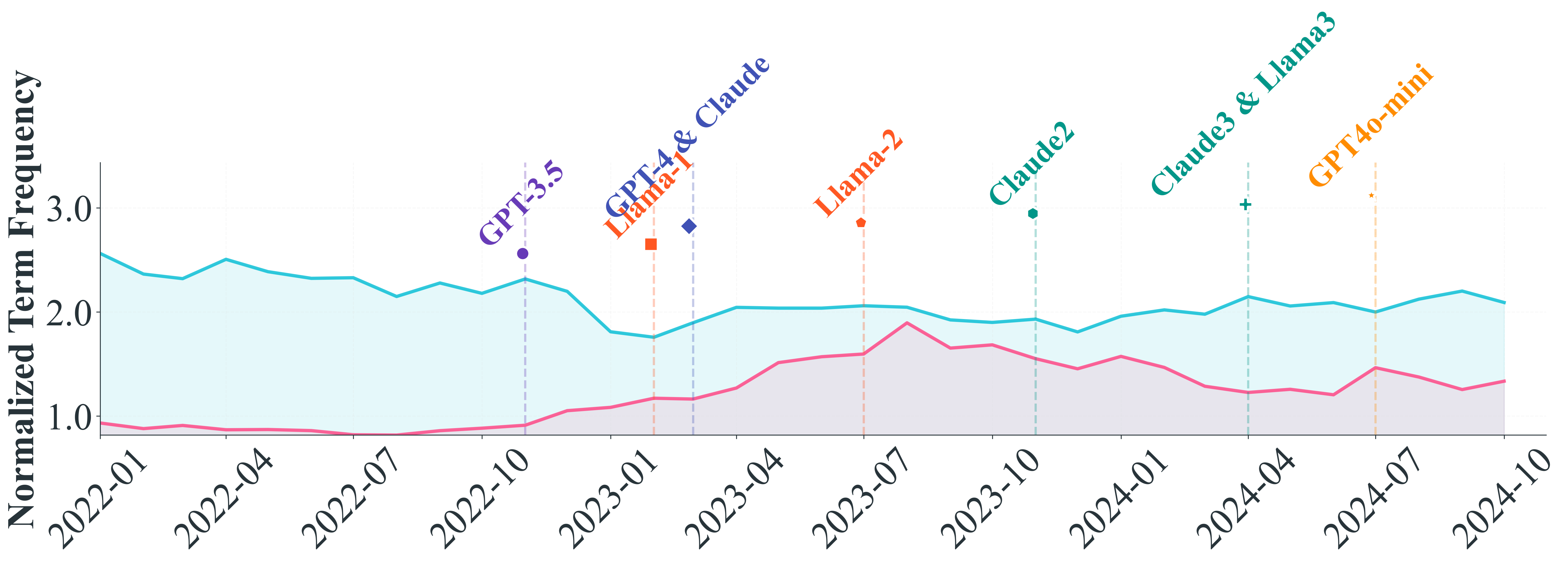}
        \caption{Word frequency trends on Quora from January 1, 2022, to October 31, 2024.}
        \label{fig:quora-words-trends-chart}
    \end{subfigure}
    \begin{subfigure}[t]{\linewidth}
        \centering
        \includegraphics[width=0.90\linewidth]
        {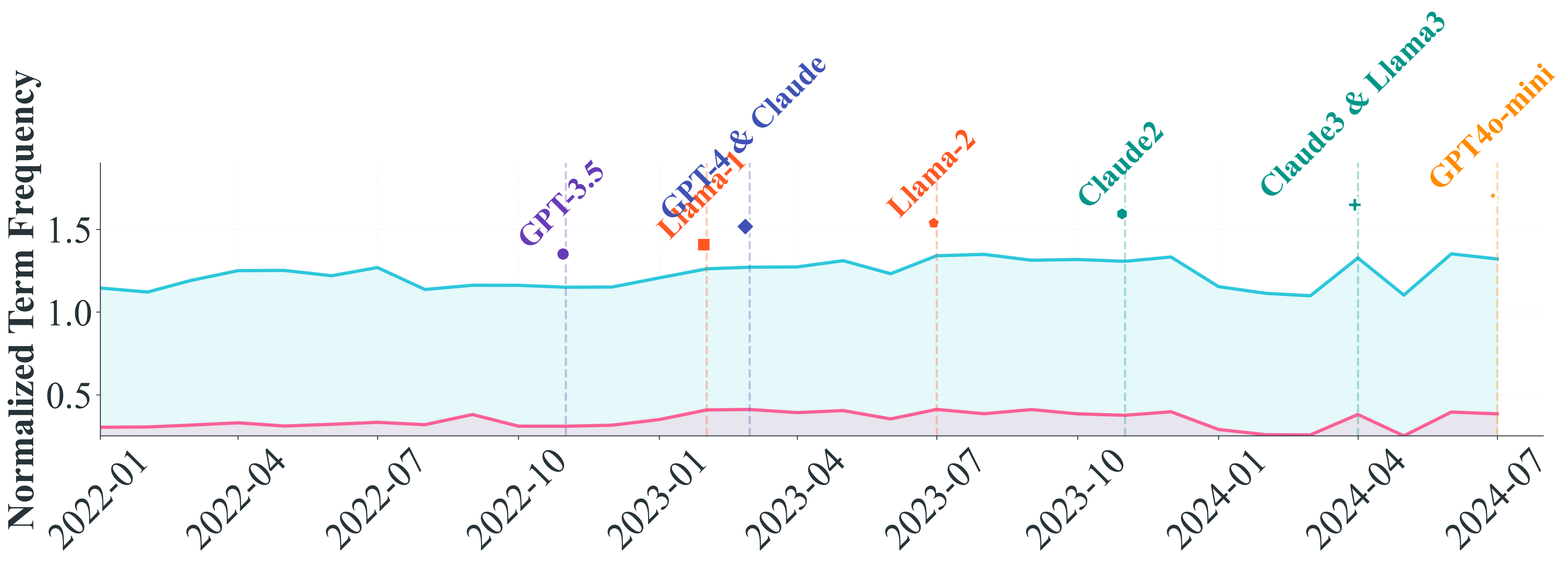}
        \caption{Word frequency trends on Reddit from January 1, 2022, to July 31, 2024.}
        \label{fig:reddit-words-trends-chart}
    \end{subfigure}
    \caption{Comparison of Medium and Quora word frequency trends: human vs. AI preferences.}     
    \label{fig:combined-words-trends}
\end{figure}

\subsection{Linguistic Analysis at Different Levels}

We explore the interpretability of the \Detector model in the case study using two methods: Integrated Gradients~\cite{sundararajan2017axiomatic}, representing a model-dependent perspective, and Shapley Value~\cite{scott2017unified}, offering a model-independent perspective.
Details of the two methods can be found in~\Cref{sec:model-interpretation-analyze-methods}. 

\mypara{Word-Level Analysis}
In the case study of Reddit (refer to ~\Cref{Reddit_intergrate_grdients,Reddit_word_level_shaplay}), words like \highlight{``and''}, \highlight{``think''} and \highlight{``I''} have the highest Integrated Gradients and Shapley Values, which lead model to classify texts as human-written.
Meanwhile, model-specific analysis shows the words \highlight{``think''}, \highlight{``can''}, and \highlight{``Online''} have the lowest scores, leading to AI-generated prediction.
From these observations, we note that specifying clear word-level patterns between two class is challenging because certain words, like \highlight{``think''}, contribute significantly to both classifications. 
This overlap suggests that word importance is highly context-dependent.
Similar challenges are also observed on Medium and Quora (\Cref{Medium_intergrate_grdients,Medium_word_level_shaplay,Quora_word_level_shaplay,Quora_intergrate_grdients}).

Given this difficulty, we then turn to a different approach: a statistical analysis of high-frequency adjectives, conjunctions, and adverbs (details provided in~\Cref{sec:collection-high-frequency-wrods}). 
These high-frequency terms are then classified into human-preferred and AI-preferred vocabularies. 
We then track the trends of these lexical items on \Dataset.

As shown in~\Cref{fig:medium-words-trends-chart,fig:quora-words-trends-chart}, the NTF of AI-preferred vocabulary on the Medium and Quora is closely aligned with the development of LLMs. 
Following the release of LLMs such as GPT, Llama, and the Claude series, the NTF of human-preferred vocabulary has gradually declined. 
Meanwhile, AI-preferred vocabulary shows an increase.
These results reflect an increasing usage of LLMs for content generation by Medium and Quora platform users.
In contrast, the trends on Reddit show some differences. 
From 2022 to 2024, the NTF of human-preferred vocabulary always remains high, while the AI-preferred vocabulary consistently remains low. 
This indicates that Reddit users rely less on LLMs to produce content.
From above, we obverse that word frequency changes closely align with the AAR trends in~\Cref{fig:overall_combined_comparison}.

\mypara{Sentence-Level Analysis}
We also conduct a sentence-level analysis using Shapley values, as Integrated Gradients are only suitable for word-level.
From the case studies of Medium, Quora, and Reddit (shown in~\Cref{Medium_sentence_level_shaplay,Quora_sentence_level_shaplay,Reddit_Sentecet_Level_Shaplay}), we observe that AIGTs are characterized by their objective and standardized structures, typically beginning with a noun or pronoun and following a verb-object pattern, like \highlight{``Online bullying...contributes...feelings...''}. 
In contrast, HWTs often contain flexible sentence structures and informal expressions, as illustrated by  \highlight{``That being said, why not both?''} and \highlight{``Why can't we restore...''}.
In summary, the results suggest that sentence-level patterns provide more distinctive characteristics for distinguishing AIGTs and HWTs, as LLMs may usually follow a standardized pattern to generate texts.

\begin{figure}[h!]
    \centering
    \includegraphics[width=0.85\linewidth]{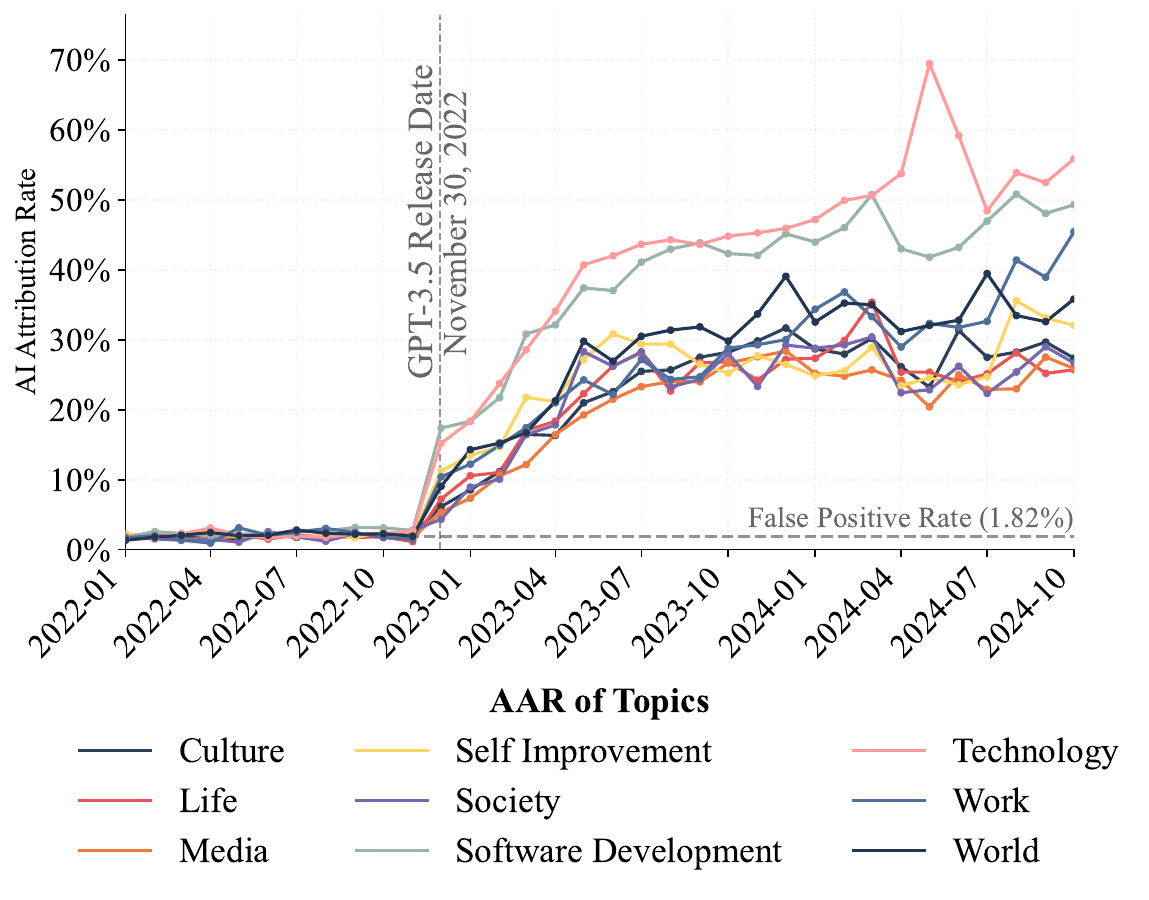}
    \caption{AAR trends across different topics.}
    \label{fig:topic}
\end{figure}

\subsection{Multidimensional Analysis of Posts}

We analyze posts on social media from multi-dimensions to find the characteristics between posts predicted as AIGTs and those classified as HWTs, including topic, engagement, and author analysis.

\mypara{Topic Analysis}
Classifying topics on platforms like Quora and Reddit is challenging due to their wide range.
Therefore, we focus our analysis on $9$ major topics listed on the Medium~\cite{MediumExploreTopics}, examining them from a temporal perspective. The proportion of topics is shown in~\Cref{fig:stacked-area-chart}.

\Cref{fig:topic} shows the trends of AAR across different topics. 
We observe a rapid increase in AAR for all topics following the release of GPT-3.5 in December 2022, indicating that the popularity of LLMs has impacted all topics on Medium.
Besides, the AAR for ``Technology'' and ``Software Development'' remains consistently higher than other topics from December 2022 to October 2024, ranking respectively first and second.
One possible reason is that people in the technology field are more likely to know about LLMs and frequently interact with them, leading to a higher AAR.

\begin{table}[H]
\centering
\small
\begin{tabular}{@{}c c c@{}}
\toprule
Follower Group  & \makecell{Mean Likes \\ (AIGTs / HWTs)} & \makecell{Mean Comments \\ (AIGTs / HWTs)} \\
\midrule 
0-1K     & $49.48 / 79.39$  & $3.18 / 5.68$  \\ 
1-5K     & $111.50 / 191.61$  & $5.11 / 9.09$  \\
\textgreater 5K  & $126.94 / 211.92$  & $5.56 / 8.25$  \\
\bottomrule 
\end{tabular}
\caption{Engagement statistics on Medium for different follower groups, comparing AIGTs and HWTs.}
\label{tab:engagement_analysis}   
\end{table}

\vspace{-2ex} 

\begin{figure}[ht]
    \centering
    \includegraphics[width=0.32\textwidth]{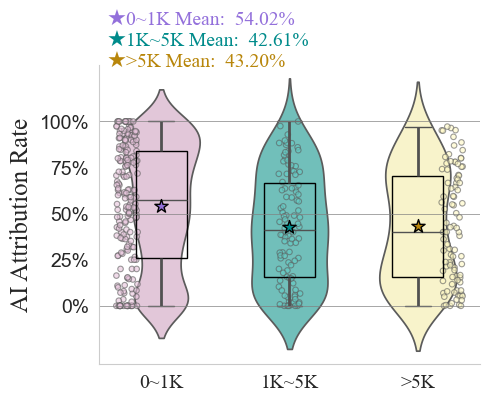}
    \caption{AAR distribution among follower groups.}
    \label{fig:followers}
\end{figure}

\mypara{Engagement Analysis}
To understand how user engagement differs between articles predicted to be AIGTs or HWTs, we analyze the number of ``Likes'' (known as ``Claps'' on Medium) and ``Comments'' in Medium blogs.
To ensure balanced comparisons, we randomly select $16,600$ blogs with a 1:1 class ratio. 
Mann-Whitney U tests reveal statistically significant differences in the number of ``Likes'' and ``Comments'' between the two classes ($p < 0.05$).

As shown in \Cref{fig:claps}, the predicted-AIGTs receive fewer ``Likes'' on average than predicted-HWTs, with mean values of $69.15$ and $127.59$, respectively.
And predicted-AIGTs exhibit a higher frequency of low ``Likes'' counts.
\Cref{fig:comments} shows that predicted-AIGTs receive fewer ``Comments'' on average compared to predicted-HWTs, with mean values of $4.16$ and $7.38$, respectively.
We further investigate the mean values of Likes and Comments for authors with different numbers of followers and \Cref{tab:engagement_analysis} indicates that, across all follower count groups, AIGTs receive significantly fewer Likes and Comments compared to HWTs.

To summarize, predicted-HWTs obtain more ``Likes'' and ``Comments'', which indicates that users in Medium are generally more willing to engage with human-written content.
However, the relatively small gap between the two suggests that AI-generated content appeals to users.

\mypara{Author Analysis}
On Medium, we randomly select $1,000$ authors from the predicted-AIGTs group who have published at least ten articles. 
We collect and detect all of their published articles to determine if they are AI-generated, aiming to explore the potential relationship between an author's follower count and their usage of AI-generated content.

As shown in~\Cref{fig:followers}, we divide these authors into three groups based on their follower count.
Among the groups, those with $1,000$ or fewer followers exhibit a stronger concentration in the high AAR range ($\ge75.00\%$).
This group also achieves the highest mean AAR at $54.02\%$. 
From the overall distribution, as the follower number increases, the AAR gradually shifts toward the lower range ($\le25.00\%$).
This trend may stem from more popular authors prioritizing content quality, while less-followed authors rely on LLMs to boost efficiency.

Furthermore, \Cref{fig:timeline_early_author_aigts} illustrates the publication timeline of the first articles detected as AIGTs from these authors.
It can be observed that there is a significant increase in such publications during the month GPT-3.5 is released, followed by a relatively stable trend in subsequent months.

\section{Conclusion}

In this paper, we collect a large-scale dataset, \Dataset, encompassing multiple platforms and diverse time periods, providing the first comprehensive quantification and analysis of AIGTs on online social media.
We construct \Bench, an AIGT detection benchmark integrating diverse LLMs, to identify the most effective detector, \Detector.
We then perform temporal tracking analyses, highlighting distinct trends in AAR that are shaped by platform-specific characteristics and the increasing adoption of LLMs.
Finally, our analysis uncovers critical differences between AIGTs and HWTs across linguistic patterns, topical features, engagement levels, and the follower distribution of authors.
Our findings offer valuable perspectives into the evolving dynamics of AIGTs on social media.

\section*{Ethical Statement}

We emphasize that the purpose of this research is not to expose or criticize specific platforms or users for employing AIGTs nor to interfere with legitimate content-creation activities.
Instead, our goal is to provide valuable insights through scientific analysis to aid the research community and the public to better understand the current state and trends of generative AI usage on social media.
All data used in our paper is publicly available, and we do not collect and monitor any private information.

\section*{Limitations}

In this paper, we conduct long-term quantification of AIGTs on $3$ commonly used social media platforms, but there are still some limitations:
\begin{enumerate}
    \item \textbf{Limited coverage of LLMs:} \Bench includes only $12$ LLMs and does not cover all LLMs released across different time periods. We only included models released after November 2022. This decision was made because our study specifically focuses on more powerful models, such as ChatGPT, which may lead to misclassifications for earlier models. In addition, the data set shows distributional bias favoring the GPT series $42.9\%$ and the Llama series $39.05\%$ models. While current AIGT detectors can generalize to unseen LLMs to some extent~\cite{li2024mage}, these coverage limitations may introduce slight errors and pose potential impacts on the accuracy of some results. However, these biases are unlikely to significantly impact the analysis results, as these models are also the most widely used in real-world applications.
    \item \textbf{Lack of analysis on multilingual platforms:} Our research focuses on English-dominated social media platforms. Therefore, the applicability of our findings is restricted to these specific platforms and language contexts. Since data collection is a long-term process, we plan to gradually expand to multilingual environments and more platforms in future research to improve the universality of the conclusions.
    \item \textbf{Insufficient dimensions of analysis across platforms:} We conduct an in-depth analysis of the three dimensions of topic, engagement, and author on the Medium platform, but we are unable to conduct similar multi-dimensional research on Quora and Reddit. This is mainly due to the differences in data collection methods and the difficulty of different platforms. If richer data from these platforms becomes available in the future, we will supplement and enhance the analysis.
\end{enumerate}

\section*{Acknowledgement}
We would like to thank all anonymous reviewers, Area Chair, and Program Chair for their insightful comments and constructive suggestions. We are also grateful to Dr. Yun Shen for his valuable guidance and feedback in this work.

\clearpage
\bibliographystyle{plainnat}
\bibliography{ref}

\begin{thebibliography}{72}
\providecommand{\natexlab}[1]{#1}
\providecommand{\url}[1]{\texttt{#1}}
\expandafter\ifx\csname urlstyle\endcsname\relax
  \providecommand{\doi}[1]{doi: #1}\else
  \providecommand{\doi}{doi: \begingroup \urlstyle{rm}\Url}\fi

\bibitem[Aaditya(2024)]{Llama3-OpenBioLLM-8B}
Aaditya.
\newblock Llama 3 - openbiollm - 8b model, 2024.
\newblock URL \url{https://huggingface.co/aaditya/Llama3-OpenBioLLM-8B}.
\newblock Accessed: 2025-02-15.

\bibitem[{Adam D'Angelo, 2023}(2024)]{quora_blog_poe}
{Adam D'Angelo, 2023}.
\newblock Poe ai introduction, 2024.
\newblock Available at: \url{https://quorablog.quora.com/Poe-1} [Accessed: 2024-12-05].

\bibitem[{Anthropic}(2024)]{anthropic_site}
{Anthropic}.
\newblock Anthropic official website, 2024.
\newblock URL \url{https://www.anthropic.com/}.
\newblock Accessed: 2024-11-04.

\bibitem[Bhattacharjee et~al.(2023)Bhattacharjee, Kumarage, Moraffah, and Liu]{bhattacharjee2023conda}
Amrita Bhattacharjee, Tharindu Kumarage, Raha Moraffah, and Huan Liu.
\newblock Conda: Contrastive domain adaptation for ai-generated text detection.
\newblock In \emph{{Annual Meeting of the Association for Computational Linguistics and International Joint Conference on Natural Language Processing (ACL/IJCNLP)}}, pages 598--610. ACL, 2023.

\bibitem[BigModel(2024)]{GLM4API}
BigModel.
\newblock Glm-4 api documentation, 2024.
\newblock URL \url{https://open.bigmodel.cn/dev/api/normal-model/glm-4}.
\newblock Accessed: 2025-02-15.

\bibitem[Briesch et~al.(2023)Briesch, Sobania, and Rothlauf]{briesch2023large}
Martin Briesch, Dominik Sobania, and Franz Rothlauf.
\newblock Large language models suffer from their own output: An analysis of the self-consuming training loop.
\newblock \emph{CoRR}, abs/2311.16822, 2023.

\bibitem[Chiang et~al.(2023)Chiang, Li, Lin, Sheng, Wu, Zhang, Zheng, Zhuang, Zhuang, Gonzalez, et~al.]{chiang2023vicuna}
Wei-Lin Chiang, Zhuohan Li, Zi~Lin, Ying Sheng, Zhanghao Wu, Hao Zhang, Lianmin Zheng, Siyuan Zhuang, Yonghao Zhuang, Joseph~E Gonzalez, et~al.
\newblock Vicuna: An open-source chatbot impressing gpt-4 with 90\%* chatgpt quality.
\newblock \emph{See https://vicuna. lmsys. org (accessed 14 April 2023)}, 2\penalty0 (3):\penalty0 6, 2023.

\bibitem[Computations(2024)]{Dolphin3.0-Llama3.1-8B}
Cognitive Computations.
\newblock Dolphin 3.0 - llama 3.1 - 8b model, 2024.
\newblock URL \url{https://huggingface.co/cognitivecomputations/Dolphin3.0-Llama3.1-8B}.
\newblock Accessed: 2025-02-15.

\bibitem[DeepMind(2024)]{GeminiFlash}
DeepMind.
\newblock Gemini flash, 2024.
\newblock URL \url{https://deepmind.google/technologies/gemini/flash/}.
\newblock Accessed: 2025-02-15.

\bibitem[Dubey et~al.(2024)Dubey, Jauhri, Pandey, Kadian, Al-Dahle, Letman, Mathur, Schelten, Yang, Fan, et~al.]{dubey2024llama}
Abhimanyu Dubey, Abhinav Jauhri, Abhinav Pandey, Abhishek Kadian, Ahmad Al-Dahle, Aiesha Letman, Akhil Mathur, Alan Schelten, Amy Yang, Angela Fan, et~al.
\newblock The llama 3 herd of models.
\newblock \emph{arXiv preprint arXiv:2407.21783}, 2024.

\bibitem[Fraser et~al.(2025)Fraser, Dawkins, and Kiritchenko]{fraser2024detecting}
Kathleen~C. Fraser, Hillary Dawkins, and Svetlana Kiritchenko.
\newblock Detecting ai-generated text: Factors influencing detectability with current methods.
\newblock \emph{J. Artif. Intell. Res.}, 82:\penalty0 2233--2278, 2025.

\bibitem[Gehrmann et~al.(2019)Gehrmann, Strobelt, and Rush]{gehrmann2019gltr}
Sebastian Gehrmann, Hendrik Strobelt, and Alexander~M. Rush.
\newblock {GLTR:} statistical detection and visualization of generated text.
\newblock In \emph{{Annual Meeting of the Association for Computational Linguistics (ACL)}}, pages 111--116. Association for Computational Linguistics, 2019.

\bibitem[{GPTZero}(2024)]{gptzero}
{GPTZero}.
\newblock Gptzero, 2024.
\newblock URL \url{https://gptzero.me/}.
\newblock Accessed: 2024-11-04.

\bibitem[Gruda(2024)]{gruda2024three}
Dritjon Gruda.
\newblock Three ways chatgpt helps me in my academic writing.
\newblock \emph{Nature}, 2024.
\newblock \doi{10.1038/d41586-024-01042-3}.
\newblock URL \url{https://doi.org/10.1038/d41586-024-01042-3}.
\newblock Advance online publication.

\bibitem[Guo et~al.(2023)Guo, Zhang, Wang, Jiang, Nie, Ding, Yue, and Wu]{guo2023close}
Biyang Guo, Xin Zhang, Ziyuan Wang, Minqi Jiang, Jinran Nie, Yuxuan Ding, Jianwei Yue, and Yupeng Wu.
\newblock How close is chatgpt to human experts? comparison corpus, evaluation, and detection.
\newblock \emph{CoRR}, abs/2301.07597, 2023.

\bibitem[Hanley and Durumeric(2024)]{hanley2024machine}
Hans W.~A. Hanley and Zakir Durumeric.
\newblock Machine-made media: Monitoring the mobilization of machine-generated articles on misinformation and mainstream news websites.
\newblock In \emph{Proceedings of the Eighteenth International {AAAI} Conference on Web and Social Media, {ICWSM} 2024, Buffalo, New York, USA, June 3-6, 2024}, pages 542--556. {AAAI} Press, 2024.

\bibitem[He et~al.(2024)He, Shen, Chen, Backes, and Zhang]{he2023mgtbench}
Xinlei He, Xinyue Shen, Zeyuan Chen, Michael Backes, and Yang Zhang.
\newblock Mgtbench: Benchmarking machine-generated text detection.
\newblock In \emph{{ACM Conference on Computer and Communications Security (CCS)}}, pages 2251--2265. {ACM}, 2024.

\bibitem[He et~al.(2025)He, Xu, Han, Wang, Zhao, Shen, Lin, Zhao, Li, Yang, Ji, Li, Zhu, Wang, Zheng, Zhu, Li, He, Wang, Hu, Wang, Sun, Yao, Qin, Chen, Zhao, Li, Huang, and Feng]{He2025AISecuritySurvey}
Xinlei He, Guowen Xu, Xingshuo Han, Qian Wang, Lingchen Zhao, Chao Shen, Chenhao Lin, Zhengyu Zhao, Qian Li, Le~Yang, Shouling Ji, Shaofeng Li, Haojin Zhu, Zhibo Wang, Rui Zheng, Tianqing Zhu, Qi~Li, Chaoxiang He, Qifan Wang, Hongsheng Hu, Shuo Wang, Shi-Feng Sun, Hongwei Yao, Zhan Qin, Kai Chen, Yue Zhao, Hongwei Li, Xinyi Huang, and Dengguo Feng.
\newblock Artificial intelligence security and privacy: a survey.
\newblock \emph{Science China Information Sciences}, 2025.

\bibitem[Heralax(2025)]{heralax_mannerstral_dataset}
Heralax.
\newblock Mannerstral-dataset, 2025.
\newblock URL \url{https://huggingface.co/datasets/Heralax/Mannerstral-dataset?row=2}.

\bibitem[Honnibal et~al.(2020)Honnibal, Montani, Van~Landeghem, and Boyd]{Honnibal_spaCy_Industrial-strength_Natural_2020}
Matthew Honnibal, Ines Montani, Sofie Van~Landeghem, and Adriane Boyd.
\newblock {spaCy: Industrial-strength Natural Language Processing in Python}.
\newblock 2020.
\newblock \doi{10.5281/zenodo.1212303}.

\bibitem[Ippolito et~al.(2020)Ippolito, Duckworth, Callison{-}Burch, and Eck]{ippolito2019automatic}
Daphne Ippolito, Daniel Duckworth, Chris Callison{-}Burch, and Douglas Eck.
\newblock Automatic detection of generated text is easiest when humans are fooled.
\newblock In \emph{{Annual Meeting of the Association for Computational Linguistics (ACL)}}, pages 1808--1822. ACL, 2020.

\bibitem[Jiang et~al.(2024)Jiang, Sablayrolles, Roux, Mensch, Savary, Bamford, Chaplot, Casas, Hanna, Bressand, et~al.]{jiang2024mixtral}
Albert~Q Jiang, Alexandre Sablayrolles, Antoine Roux, Arthur Mensch, Blanche Savary, Chris Bamford, Devendra~Singh Chaplot, Diego de~las Casas, Emma~Bou Hanna, Florian Bressand, et~al.
\newblock Mixtral of experts.
\newblock \emph{arXiv preprint arXiv:2401.04088}, 2024.

\bibitem[Kamalloo et~al.(2023)Kamalloo, Dziri, Clarke, and Rafiei]{kamalloo2023evaluating}
Ehsan Kamalloo, Nouha Dziri, Charles L.~A. Clarke, and Davood Rafiei.
\newblock Evaluating open-domain question answering in the era of large language models.
\newblock In \emph{{Annual Meeting of the Association for Computational Linguistics (ACL)}}, pages 5591--5606. ACL, 2023.

\bibitem[Kokhlikyan et~al.(2020)Kokhlikyan, Miglani, Martin, Wang, Alsallakh, Reynolds, Melnikov, Kliushkina, Araya, Yan, and Reblitz{-}Richardson]{kokhlikyan2020captum}
Narine Kokhlikyan, Vivek Miglani, Miguel Martin, Edward Wang, Bilal Alsallakh, Jonathan Reynolds, Alexander Melnikov, Natalia Kliushkina, Carlos Araya, Siqi Yan, and Orion Reblitz{-}Richardson.
\newblock Captum: {A} unified and generic model interpretability library for pytorch.
\newblock \emph{CoRR}, abs/2009.07896, 2020.

\bibitem[Li et~al.(2024)Li, Li, Cui, Bi, Wang, Wang, Yang, Shi, and Zhang]{li2024mage}
Yafu Li, Qintong Li, Leyang Cui, Wei Bi, Zhilin Wang, Longyue Wang, Linyi Yang, Shuming Shi, and Yue Zhang.
\newblock {MAGE:} machine-generated text detection in the wild.
\newblock In \emph{{Annual Meeting of the Association for Computational Linguistics (ACL)}}, pages 36--53. ACL, 2024.

\bibitem[Liu et~al.(2024{\natexlab{a}})Liu, Feng, Xue, Wang, Wu, Lu, Zhao, Deng, Zhang, Ruan, et~al.]{liu2024deepseek}
Aixin Liu, Bei Feng, Bing Xue, Bingxuan Wang, Bochao Wu, Chengda Lu, Chenggang Zhao, Chengqi Deng, Chenyu Zhang, Chong Ruan, et~al.
\newblock Deepseek-v3 technical report.
\newblock \emph{arXiv preprint arXiv:2412.19437}, 2024{\natexlab{a}}.

\bibitem[Liu et~al.(2023)Liu, Zhang, Wang, Pu, Lan, and Shen]{liu2023coco}
Xiaoming Liu, Zhaohan Zhang, Yichen Wang, Hang Pu, Yu~Lan, and Chao Shen.
\newblock Coco: Coherence-enhanced machine-generated text detection under low resource with contrastive learning.
\newblock In \emph{{Conference on Empirical Methods in Natural Language Processing (EMNLP)}}, pages 16167--16188. ACL, 2023.

\bibitem[Liu et~al.(2019)Liu, Ott, Goyal, Du, Joshi, Chen, Levy, Lewis, Zettlemoyer, and Stoyanov]{liu2019roberta}
Yinhan Liu, Myle Ott, Naman Goyal, Jingfei Du, Mandar Joshi, Danqi Chen, Omer Levy, Mike Lewis, Luke Zettlemoyer, and Veselin Stoyanov.
\newblock Roberta: {A} robustly optimized {BERT} pretraining approach.
\newblock \emph{CoRR}, abs/1907.11692, 2019.

\bibitem[Liu et~al.(2024{\natexlab{b}})Liu, Sun, He, and Huang]{liu2024quantized}
Yule Liu, Zhen Sun, Xinlei He, and Xinyi Huang.
\newblock Quantized delta weight is safety keeper.
\newblock \emph{CoRR}, abs/2411.19530, 2024{\natexlab{b}}.

\bibitem[Liu et~al.(2025)Liu, Zhong, Liao, Sun, Zheng, Wei, Gong, Tong, Chen, Zhang, and He]{liu2024generalizationabilitymachinegeneratedtext}
Yule Liu, Zhiyuan Zhong, Yifan Liao, Zhen Sun, Jingyi Zheng, Jiaheng Wei, Qingyuan Gong, Fenghua Tong, Yang Chen, Yang Zhang, and Xinlei He.
\newblock {On the Generalization and Adaptation Ability of Machine-Generated Text Detectors in Academic Writing}.
\newblock In \emph{{Proceedings of the 31st ACM International Conference on Knowledge Discovery and Data Mining (KDD), Volume 2}}. ACM, 2025.

\bibitem[Liu et~al.(2024{\natexlab{c}})Liu, Yao, Li, and Luo]{liu2023detectability}
Zeyan Liu, Zijun Yao, Fengjun Li, and Bo~Luo.
\newblock On the detectability of chatgpt content: Benchmarking, methodology, and evaluation through the lens of academic writing.
\newblock In \emph{{ACM SIGSAC Conference on Computer and Communications Security (CCS)}}, pages 2236--2250. {ACM}, 2024{\natexlab{c}}.

\bibitem[Lundberg and Lee(2017)]{scott2017unified}
Scott~M. Lundberg and Su{-}In Lee.
\newblock A unified approach to interpreting model predictions.
\newblock In \emph{Advances in neural information processing systems}, pages 4765--4774, 2017.

\bibitem[Macko et~al.(2024)Macko, Kopal, M{\'{o}}ro, and Srba]{macko2024multisocial}
Dominik Macko, Jakub Kopal, R{\'{o}}bert M{\'{o}}ro, and Ivan Srba.
\newblock Multisocial: Multilingual benchmark of machine-generated text detection of social-media texts.
\newblock \emph{CoRR}, abs/2406.12549, 2024.

\bibitem[Magpie-Align(2025{\natexlab{a}})]{magpie_reasoning_v1_150k_cot_qwq}
Magpie-Align.
\newblock Magpie-reasoning-v1-150k-cot-qwq, 2025{\natexlab{a}}.
\newblock URL \url{https://huggingface.co/datasets/Magpie-Align/Magpie-Reasoning-V1-150K-CoT-QwQ}.

\bibitem[Magpie-Align(2025{\natexlab{b}})]{magpie_reasoning_v2_250k_cot_deepseek_r1_llama_70b}
Magpie-Align.
\newblock Magpie-reasoning-v2-250k-cot-deepseek-r1-llama-70b, 2025{\natexlab{b}}.
\newblock URL \url{https://huggingface.co/datasets/Magpie-Align/Magpie-Reasoning-V2-250K-CoT-Deepseek-R1-Llama-70B}.

\bibitem[Medium(2024)]{MediumExploreTopics}
Medium.
\newblock Explore topics on medium, 2024.
\newblock URL \url{https://medium.com/explore-topics}.
\newblock Accessed: 2025-02-15.

\bibitem[{Medium}(2024)]{medium}
{Medium}.
\newblock Medium, 2024.
\newblock URL \url{https://medium.com/}.
\newblock Accessed: 2024-11-04.

\bibitem[{Meta AI}(2024)]{meta2024llamareport}
{Meta AI}.
\newblock With 10x growth since 2023, llama is the leading engine of ai innovation, 2024.
\newblock URL \url{https://ai.meta.com/blog/llama-usage-doubled-may-through-july-2024/}.
\newblock Accessed: 2024-11-04.

\bibitem[Mitchell et~al.(2023)Mitchell, Lee, Khazatsky, Manning, and Finn]{mitchell2023detectgpt}
Eric Mitchell, Yoonho Lee, Alexander Khazatsky, Christopher~D Manning, and Chelsea Finn.
\newblock Detectgpt: Zero-shot machine-generated text detection using probability curvature.
\newblock In \emph{International Conference on Machine Learning (ICML)}, pages 24950--24962. PMLR, 2023.

\bibitem[{Moonshot}(2024)]{moonshot_llm}
{Moonshot}.
\newblock Mootshot llm, 2024.
\newblock URL \url{https://kimi.moonshot.cn/}.
\newblock Accessed: 2024-11-04.

\bibitem[OdiaGenAI(2025)]{odiagenai_roleplay_english}
OdiaGenAI.
\newblock Roleplay-english, 2025.
\newblock URL \url{https://huggingface.co/datasets/OdiaGenAI/roleplay_english?row=24}.

\bibitem[{OpenAI}(2022)]{openai_chatgpt}
{OpenAI}.
\newblock Introducing chatgpt, 2022.
\newblock URL \url{https://openai.com/index/chatgpt/}.
\newblock Accessed: 2024-11-04.

\bibitem[OpenAI(2023)]{openai2023gpt4}
OpenAI.
\newblock Gpt-4 technical report, 2023.

\bibitem[{OpenAI}(2024)]{openai_gpt4o_mini}
{OpenAI}.
\newblock Gpt-4o mini: Advancing cost-efficient intelligence, 2024.
\newblock URL \url{https://openai.com/index/gpt-4o-mini-advancing-cost-efficient-intelligence/}.
\newblock Accessed: 2024-11-04.

\bibitem[OpenGVLab(2024)]{InternVL2_5-8B}
OpenGVLab.
\newblock Internvl2.5-8b model, 2024.
\newblock URL \url{https://huggingface.co/OpenGVLab/InternVL2_5-8B}.
\newblock Accessed: 2025-02-15.

\bibitem[OpenGVLab(2025)]{openGVLab_unternvl_as_1b_caption}
OpenGVLab.
\newblock Internvl-sa-1b-caption, 2025.
\newblock URL \url{https://huggingface.co/datasets/OpenGVLab/InternVL-SA-1B-Caption?row=0}.

\bibitem[PJMixers-Dev(2025)]{customsharegpt}
PJMixers-Dev.
\newblock camel-ai\_chemistry-gemini-2.0-flash-thinking-exp-1219-customsharegpt, 2025.
\newblock URL \url{https://huggingface.co/datasets/PJMixers-Dev/camel-ai_chemistry-gemini-2.0-flash-thinking-exp-1219-CustomShareGPT}.

\bibitem[{Quora}(2024)]{quora}
{Quora}.
\newblock Quora, 2024.
\newblock URL \url{https://www.quora.com/}.
\newblock Accessed: 2024-11-04.

\bibitem[Radford et~al.(2019)Radford, Wu, Child, Luan, Amodei, Sutskever, et~al.]{radford2019language}
Alec Radford, Jeffrey Wu, Rewon Child, David Luan, Dario Amodei, Ilya Sutskever, et~al.
\newblock Language models are unsupervised multitask learners.
\newblock \emph{OpenAI blog}, 1\penalty0 (8):\penalty0 9, 2019.

\bibitem[{Reddit}(2024)]{reddit}
{Reddit}.
\newblock Reddit, 2024.
\newblock URL \url{https://www.reddit.com/}.
\newblock Accessed: 2024-11-04.

\bibitem[Shapley(1953)]{shapley1953value}
Lloyd~S Shapley.
\newblock A value for n-person games.
\newblock \emph{Contribution to the Theory of Games}, 2, 1953.

\bibitem[Shen et~al.(2025)Shen, Wu, Qu, Backes, Zannettou, and Zhang]{SWQBZZ25}
Xinyue Shen, Yixin Wu, Yiting Qu, Michael Backes, Savvas Zannettou, and Yang Zhang.
\newblock {HateBench: Benchmarking Hate Speech Detectors on LLM-Generated Content and Hate Campaigns}.
\newblock In \emph{{USENIX Security Symposium (USENIX Security)}}. USENIX, 2025.

\bibitem[Shi et~al.(2024)Shi, Sheng, Cao, Mi, Hu, and Wang]{shi2024ten}
Yuhui Shi, Qiang Sheng, Juan Cao, Hao Mi, Beizhe Hu, and Danding Wang.
\newblock Ten words only still help: Improving black-box ai-generated text detection via proxy-guided efficient re-sampling.
\newblock In \emph{{International Joint Conferences on Artifical Intelligence (IJCAI)}}, pages 494--502. ijcai.org, 2024.

\bibitem[Solaiman et~al.(2019)Solaiman, Brundage, Clark, Askell, Herbert{-}Voss, Wu, Radford, and Wang]{solaiman2019release}
Irene Solaiman, Miles Brundage, Jack Clark, Amanda Askell, Ariel Herbert{-}Voss, Jeff Wu, Alec Radford, and Jasmine Wang.
\newblock Release strategies and the social impacts of language models.
\newblock \emph{CoRR}, abs/1908.09203, 2019.

\bibitem[Soto et~al.(2024)Soto, Koch, Khan, Chen, Bishop, and Andrews]{soto2024few}
Rafael A.~Rivera Soto, Kailin Koch, Aleem Khan, Barry~Y. Chen, Marcus Bishop, and Nicholas Andrews.
\newblock Few-shot detection of machine-generated text using style representations.
\newblock In \emph{{International Conference on Learning Representations (ICLR)}}. OpenReview.net, 2024.

\bibitem[Su et~al.(2023)Su, Zhuo, Wang, and Nakov]{su2023detectllm}
Jinyan Su, Terry~Yue Zhuo, Di~Wang, and Preslav Nakov.
\newblock Detectllm: Leveraging log rank information for zero-shot detection of machine-generated text.
\newblock In \emph{Findings of the Association for Computational Linguistics: {EMNLP} 2023, Singapore, December 6-10, 2023}, pages 12395--12412. ACL, 2023.

\bibitem[Sun et~al.(2025)Sun, Cong, Liu, Lin, He, Chen, Han, and Huang]{sun2024peftguard}
Zhen Sun, Tianshuo Cong, Yule Liu, Chenhao Lin, Xinlei He, Rongmao Chen, Xingshuo Han, and Xinyi Huang.
\newblock {PEFTGuard: Detecting Backdoor Attacks Against Parameter-Efficient Fine-Tuning}.
\newblock In \emph{2025 IEEE Symposium on Security and Privacy (SP)}, pages 1620--1638. IEEE Computer Society, 2025.
\newblock \doi{10.1109/SP61157.2025.00161}.
\newblock URL \url{https://doi.ieeecomputersociety.org/10.1109/SP61157.2025.00161}.

\bibitem[Sundararajan et~al.(2017)Sundararajan, Taly, and Yan]{sundararajan2017axiomatic}
Mukund Sundararajan, Ankur Taly, and Qiqi Yan.
\newblock Axiomatic attribution for deep networks.
\newblock In \emph{{International Conference on Machine Learning (ICML)}}, volume~70, pages 3319--3328. {PMLR}, 2017.

\bibitem[Taori et~al.(2023)Taori, Gulrajani, Zhang, Dubois, Li, Guestrin, Liang, and Hashimoto]{taori2023alpaca}
Rohan Taori, Ishaan Gulrajani, Tianyi Zhang, Yann Dubois, Xuechen Li, Carlos Guestrin, Percy Liang, and Tatsunori~B Hashimoto.
\newblock Alpaca: A strong, replicable instruction-following model.
\newblock \emph{Stanford Center for Research on Foundation Models. https://crfm. stanford. edu/2023/03/13/alpaca. html}, 3\penalty0 (6):\penalty0 7, 2023.

\bibitem[Touvron et~al.(2023{\natexlab{a}})Touvron, Lavril, Izacard, Martinet, Lachaux, Lacroix, Rozi{\`e}re, Goyal, Hambro, Azhar, et~al.]{touvron2023llama}
Hugo Touvron, Thibaut Lavril, Gautier Izacard, Xavier Martinet, Marie-Anne Lachaux, Timoth{\'e}e Lacroix, Baptiste Rozi{\`e}re, Naman Goyal, Eric Hambro, Faisal Azhar, et~al.
\newblock Llama: Open and efficient foundation language models.
\newblock \emph{arXiv preprint arXiv:2302.13971}, 2023{\natexlab{a}}.

\bibitem[Touvron et~al.(2023{\natexlab{b}})Touvron, Martin, Stone, Albert, Almahairi, Babaei, Bashlykov, Batra, Bhargava, Bhosale, et~al.]{touvron2023llama2}
Hugo Touvron, Louis Martin, Kevin Stone, Peter Albert, Amjad Almahairi, Yasmine Babaei, Nikolay Bashlykov, Soumya Batra, Prajjwal Bhargava, Shruti Bhosale, et~al.
\newblock Llama 2: Open foundation and fine-tuned chat models.
\newblock \emph{arXiv preprint arXiv:2307.09288}, 2023{\natexlab{b}}.

\bibitem[Uchendu et~al.(2021)Uchendu, Ma, Le, Zhang, and Lee]{uchendu2021turingbench}
Adaku Uchendu, Zeyu Ma, Thai Le, Rui Zhang, and Dongwon Lee.
\newblock {TURINGBENCH:} {A} benchmark environment for turing test in the age of neural text generation.
\newblock In \emph{Findings of the Association for Computational Linguistics: {EMNLP} 2021, Virtual Event / Punta Cana, Dominican Republic, 16-20 November, 2021}, pages 2001--2016. Association for Computational Linguistics, 2021.

\bibitem[Vasilatos et~al.(2023)Vasilatos, Alam, Rahwan, Zaki, and Maniatakos]{vasilatos2023howkgpt}
Christoforos Vasilatos, Manaar Alam, Talal Rahwan, Yasir Zaki, and Michail Maniatakos.
\newblock Howkgpt: Investigating the detection of chatgpt-generated university student homework through context-aware perplexity analysis.
\newblock \emph{arXiv preprint arXiv:2305.18226}, 2023.

\bibitem[Wang et~al.(2023{\natexlab{a}})Wang, Lyu, Ji, Zhang, Yu, Shi, and Tu]{beltagy2020Longformer}
Longyue Wang, Chenyang Lyu, Tianbo Ji, Zhirui Zhang, Dian Yu, Shuming Shi, and Zhaopeng Tu.
\newblock Document-level machine translation with large language models.
\newblock In Houda Bouamor, Juan Pino, and Kalika Bali, editors, \emph{{Conference on Empirical Methods in Natural Language Processing (EMNLP)}}, pages 16646--16661. ACL, 2023{\natexlab{a}}.

\bibitem[Wang et~al.(2023{\natexlab{b}})Wang, Lyu, Ji, Zhang, Yu, Shi, and Tu]{wang2023document}
Longyue Wang, Chenyang Lyu, Tianbo Ji, Zhirui Zhang, Dian Yu, Shuming Shi, and Zhaopeng Tu.
\newblock Document-level machine translation with large language models.
\newblock In \emph{{Conference on Empirical Methods in Natural Language Processing (EMNLP)}}, pages 16646--16661. ACL, 2023{\natexlab{b}}.

\bibitem[Wang et~al.(2023{\natexlab{c}})Wang, Pan, Yan, Su, and Luan]{10221755}
Yuntao Wang, Yanghe Pan, Miao Yan, Zhou Su, and Tom~H. Luan.
\newblock A survey on chatgpt: Ai-generated contents, challenges, and solutions.
\newblock \emph{{IEEE} Open J. Comput. Soc.}, 4:\penalty0 280--302, 2023{\natexlab{c}}.

\bibitem[Xwin-LM(2024)]{Xwin-LM-13B-V0.2}
Xwin-LM.
\newblock Xwin-lm 13b v0.2 model, 2024.
\newblock URL \url{https://huggingface.co/Xwin-LM/Xwin-LM-13B-V0.2}.
\newblock Accessed: 2025-02-15.

\bibitem[Young et~al.(2024)Young, Chen, Li, Huang, Zhang, Zhang, Wang, Li, Zhu, Chen, et~al.]{young2024yi}
Alex Young, Bei Chen, Chao Li, Chengen Huang, Ge~Zhang, Guanwei Zhang, Guoyin Wang, Heng Li, Jiangcheng Zhu, Jianqun Chen, et~al.
\newblock Yi: Open foundation models by 01. ai.
\newblock \emph{arXiv preprint arXiv:2403.04652}, 2024.

\bibitem[Zellers et~al.(2019)Zellers, Holtzman, Rashkin, Bisk, Farhadi, Roesner, and Choi]{zellers2019defending}
Rowan Zellers, Ari Holtzman, Hannah Rashkin, Yonatan Bisk, Ali Farhadi, Franziska Roesner, and Yejin Choi.
\newblock Defending against neural fake news.
\newblock \emph{Advances in neural information processing systems}, 32, 2019.

\bibitem[Zhao et~al.(2023)Zhao, Zhou, Li, Tang, Wang, Hou, Min, Zhang, Zhang, Dong, et~al.]{zhao2023survey}
Wayne~Xin Zhao, Kun Zhou, Junyi Li, Tianyi Tang, Xiaolei Wang, Yupeng Hou, Yingqian Min, Beichen Zhang, Junjie Zhang, Zican Dong, et~al.
\newblock A survey of large language models.
\newblock \emph{arXiv preprint arXiv:2303.18223}, 2023.

\bibitem[Zheng et~al.(2025)Zheng, Hu, Cong, and He]{zheng2025cl}
Jingyi Zheng, Tianyi Hu, Tianshuo Cong, and Xinlei He.
\newblock Cl-attack: Textual backdoor attacks via cross-lingual triggers.
\newblock In \emph{Proceedings of the AAAI Conference on Artificial Intelligence}, volume~39, pages 26427--26435, 2025.

\bibitem[Zhou et~al.(2023)Zhou, Zhang, Luo, Parker, and Choudhury]{zhou2023synthetic}
Jiawei Zhou, Yixuan Zhang, Qianni Luo, Andrea~G. Parker, and Munmun~De Choudhury.
\newblock Synthetic lies: Understanding ai-generated misinformation and evaluating algorithmic and human solutions.
\newblock In \emph{{Annual ACM Conference on Human Factors in Computing Systems (CHI)}}, pages 436:1--436:20. {ACM}, 2023.

\end{thebibliography}

\clearpage
\appendix

\renewcommand{\thefigure}{A\arabic{figure}}
\renewcommand{\thetable}{A\arabic{table}}
\setcounter{figure}{0}
\setcounter{table}{0}

\section{Introduction of LLMs in Detector Training Dataset}
\label{sec:model_intro}

In this paper, we have selected the most representative LLMs as our detection targets:

\begin{itemize}
    \item \textbf{Llama-1 (Feb. 2023)}~\cite{touvron2023llama}, \textbf{Llama-2 (Jul. 2023)}\cite{touvron2023llama2}, and \textbf{Llama-3 (Apr. 2024)}~\cite{dubey2024llama}: The Llama series (from Llama-1 to Llama-3) launched by Meta are powerful and extremely popular open source models. This series of models enables researchers to fine-tune diverse datasets, is highly scalable, and is suitable for various research and development environments. The latest version, Llama-3, is equipped with a larger parameter size and optimized training architecture, making it perform better in text generation, context understanding, and complex task processing.
    \item \textbf{ChatGPT/GPT-3.5 Turbo (Nov. 2022)}~\cite{openai_chatgpt}: GPT-3.5, an optimized version of GPT-3 by OpenAI, was released in 2022. By incorporating a Reinforcement Learning from Human Feedback (RLHF) reward mechanism and human feedback data, GPT-3.5 achieves significant improvements in accuracy and coherence in text generation. This version includes the Text-DaVinci-003 and GPT-3.5 (or GPT-3.5 Turbo), which focuses on fluent and natural multi-turn conversations and serves as the core model for systems like ChatGPT website.
    \item \textbf{GPT4o-mini (Jul. 2024)}~\cite{openai_gpt4o_mini}: Developed by OpenAI, GPT4o-mini is a lightweight language model optimized from GPT-4o technology. This model is designed to deliver efficient language processing capabilities that are suitable for applications with lower resource requirements. It supports both text and visual input, with future plans to expand into audio and video input and output. Since its release, the GPT4o-mini has progressively replaced the GPT-3.5 Turbo as the core model on the ChatGPT website.
    \item \textbf{Claude (Mar. 2023)}~\cite{anthropic_site}, : Claude is an advanced AI assistant developed by Anthropic. It is a closed-source model designed to communicate efficiently and intuitively with users through NLP technology. Claude can understand and generate human language to assist users in completing a variety of tasks, including answering questions, writing content, and programming assistance.
    \item \textbf{Alpaca 7B (Mar. 2023)}~\cite{taori2023alpaca}: Alpaca 7B is a lightweight instruction-following model released by Stanford University, based on Meta's Llama-7B model and fine-tuned on the dataset of $52,000$ instruction-following examples. This fine-tuning markedly enhances the model's performance in understanding and executing task instructions. In evaluations of single-turn instruction-following tasks, Alpaca demonstrates performance comparable to OpenAI's Text-DaVinci-003, exhibiting high-quality responses to instructions.
    \item \textbf{Vicuna 13B (Mar. 2023)}~\cite{chiang2023vicuna}: Released by the LMSYS team, Vicuna 13B is based on Meta's Llama-13B model and trained on a large dataset of conversation data aggregated from high-quality models like GPT-3.5. The goal is to develop an open-source conversational model that approaches the quality of GPT-3.5.
    \item \textbf{Moonshot-v1 (Oct. 2023)}~\cite{moonshot_llm}: Developed by Moonshot AI, Moonshot-v1 is an advanced large language model for text generation. This model can understand and generate natural language text, manage everyday conversational exchanges, and produce structured content in various forms, such as articles, code, and summaries, across specialized domains.
    \item \textbf{Mixtral $8\times7$B (Dec. 2023)}~\cite{jiang2024mixtral}: Developed by Mistral AI, this LLM employs a Sparse Mixture of Experts (SMoE) architecture. It has demonstrated exceptional performance across multiple benchmarks, surpassing models like Llama-2 70B and GPT-3.5, especially excelling in tasks involving mathematics, code generation, and multilingual understanding.

\end{itemize}

\section{Introduction of Detectors}
\label{sec:detector_intro}

In this work, we adopt metric-based detectors from the MGTBench framework to detect AIGTs, including:
\begin{itemize}
    \item \textbf{Log-Likelihood}~\cite{solaiman2019release}: We evaluate the likelihood of text generation by computing its log-likelihood score under a specific language model. The model constructs a reference distribution based on HWTs and AIGTs to calculate the log-likelihood score of the input text. A higher score suggests a greater likelihood of the text being LLM-generated.
    \item \textbf{Rank}~\cite{gehrmann2019gltr} and \textbf{Log-Rank}~\cite{mitchell2023detectgpt}: The Rank method identifies the source of generation by analyzing the ranking of each word in the text. The model calculates the absolute ranking of each word based on context and averages all word rankings to derive an overall score. Generally, a lower score indicates that the text is more likely to be LLM-generated. Log-Rank, a variant of Rank, employs a logarithmic function when calculating each word's ranking, enhancing the detection of AIGTs.
    \item \textbf{Entropy}~\cite{gehrmann2019gltr}: The Entropy method calculates the average entropy value of each word in the text under context conditions. Studies show that AIGTs tend to have lower entropy values.
    \item \textbf{GLTR}~\cite{gehrmann2019gltr}: GLTR is a supportive tool for detecting AIGTs that use the ranking of words generated by a language model to sort the vocabulary of the text by predicted probability. Following Guo~\etal~\cite{guo2023close}, we employ the Test-2 feature to analyze the proportion of words in the top $10$, $100$, and $1000$ ranks to assess the generative nature of the text.
    \item \textbf{DetectGPT}~\cite{mitchell2023detectgpt}, \textbf{NPR}, and \textbf{LRR}~\cite{su2023detectllm}: The DetectGPT method introduces minor perturbations into the original text and observes changes in the model's log probability to detect its source. AIGTs typically reside at the local optima of the model's log probability function, whereas HWTs show greater changes in log probability after perturbation. The NPR method, similar to DetectGPT, focuses on observing significant increases in log-rank following perturbations to differentiate between AIGTs and HWTs. By combining log-likelihood and log-rank information, the LRR method captures the adaptiveness of generated texts in probability distributions while reflecting the text's ordinal preference relative to HWTs. This dual metric markedly enhances the detection accuracy.
\end{itemize}

\noindent We also consider model-based detectors, including:
\begin{itemize}
    \item \textbf{OpenAI Detector}~\cite{solaiman2019release}: This detector fine-tunes a RoBERTa~\cite{liu2019roberta} model using output data generated by the GPT-2 large, which has 1.5 billion parameters, to predict whether texts are LLM-generated.
    \item \textbf{ChatGPT Detector}~\cite{guo2023close}: Trained using the HC3 dataset, this approach employs a RoBERTa model and various training methods to distinguish between human and AIGTs. We select one that uses only the response texts to align with other detectors, following instructions described by He~\cite{he2023mgtbench}.
    \item \textbf{ConDA}~\cite{bhattacharjee2023conda}: This method enhances model discrimination of text sources in the feature space by maximizing the feature differences between generated samples and real samples. It also introduces a contrastive learning loss to improve detection accuracy.
    \item \textbf{GPTZero}~\cite{gptzero}: A tool aimed at AIGT detection that analyses the perplexity and burstiness of texts to determine their generative nature. GPTZero provides a public API interface capable of returning a confidence score indicating whether a text is LLM-generated.
    \item \textbf{CheckGPT}~\cite{liu2023detectability}: 
    The CheckGPT uses the pre-trained Roberta model to extract text features. Then, it uses LSTM to classify the text features and determine whether the text is LLM-generated or human-generated.
    \item \textbf{LM-D Detector}~\cite{ippolito2019automatic}: This approach adds an additional classification layer to a pre-trained language model (like RoBERTa) and fine-tunes it to differentiate between human-made and AIGTs. Inspired by the research of Li~\etal\cite{li2024mage}, which shows that Longformer~\cite{beltagy2020Longformer} has robust performance in detecting AIGT in out-of-domain texts, we also use the Longformer-base-4096 model to assess its performance in AIGT detection.
\end{itemize}

\section{Social Media Platforms}
\label{sec:social_media_platoforms}

To select suitable social media platforms for testing AIGT detection, we particularly consider the platform's mainstream status, the diversity of content, and their unique characteristics. 
Ultimately, we choose Reddit, Medium, and Quora as representative platforms.
\begin{itemize}
    \item Reddit~\cite{reddit} is a social discussion platform where users autonomously create and manage ``subreddit'' sections featuring diverse and rich content themes. All content on the site is categorized into different ``subreddits'' according to user interests, covering a wide range of topics from technology to social issues. We choose Reddit not only for its active user base—with around $330M$ monthly active users—but also for its vast content diversity, including millions of subreddit topics, allowing it to cover a variety of discussion scenarios.
    \item Medium~\cite{medium} is an American online publishing platform developed by Evan Williams and launched in August 2012. It centers on high-quality original articles and blog content and exemplifies social journalism, known for its content's depth, length, and professionalism.
    \item Quora~\cite{quora} is a platform to gain and share knowledge. It enables users to ask questions and connect with people who provide unique insights or quality answers. Users can pose questions and receive answers from other users on topics ranging from daily life to highly specialized academic, technical, and professional queries.
\end{itemize}
We have selected these $3$ platforms because their main functionalities closely align with common use cases for LLMs, such as writing and question-answering. 
Based on this, we hypothesize that there may be instances where users utilize LLMs to generate content on these platforms.

\section{Introduction of Open Source Datasets for Training Detectors}
\label{sec:training_detector_intro}

We consider $6$ publicly available AIGT datasets and $5$ common supervised finetuning datasets as one part of \Bench. 

\begin{itemize}
    \item The \textbf{MGT-Academic} dataset~\cite{liu2024generalizationabilitymachinegeneratedtext}, assembled from textual sources such as Wikipedia, arXiv, and Project Gutenberg, covers STEM, Social Sciences, and Humanities. It is generated by various LLMs, including Llama3, GPT-3.5 Turbo, Moonshot, and Mixtral $8\times7B$, forming a comprehensive AIGT dataset.
    \item The \textbf{Coco-GPT3.5} dataset~\cite{liu2023coco}, produced using OpenAI's text-davinci-0035 model, incorporates entire newspaper articles from December 2022 to February 2023, reflecting the latest content of that period.
    \item The \textbf{GPABench2} dataset~\cite{liu2023detectability}, based on the GPT-3.5 Turbo model, focuses on $3$ LLM-generated tasks: GPT-written, GPT-completed, and GPT-polished, all based on academic abstracts. Due to the extensive amount of text generated by GPT-3.5 Turbo, we sampled around $100M$ tokens from this dataset for compilation.
    \item The \textbf{LWD} dataset~\cite{soto2024few} involves texts generated by Llama-2, GPT-4, and ChatGPT. Researchers designed specific prompts to ``write an Amazon review in the style of the author of the following review: <human review>'', where each prompt incorporates a real human-written Amazon review as a stylistic reference.
    \item The \textbf{HC3} dataset~\cite{guo2023close}, collected by researchers, comprises nearly 40,000 questions and their answers from human experts and ChatGPT, covering a broad range of fields including open-domain, computer science, finance, medicine, law, and psychology.
    \item The \textbf{AIGT} dataset~\cite{shi2024ten} samples human-generated content and content from seven popular open-source or API-driven LLMs, applied in real-world scenarios such as low-quality content generation, news fabrication, and student cheating. Due to the markedly lesser capabilities of GPT-2 XL and GPT-J compared to GPT-3.5, these models were not included.
    \item Given that high-quality Supervised Finetuning (SFT) datasets are frequently used for finetuning LLMs, and considering the lack of Claude and GPT-4 model-related content in the AIGT detection datasets, we also incorporate four SFT datasets with instruction-following features: \textbf{Claude2-Alpaca}\footnote{\url{https://github.com/Lichang-Chen/claude2-alpaca}.}, \textbf{Claude-3-Opus-Claude-3.5-Sonnet-9k}\footnote{\url{https://huggingface.co/datasets/QuietImpostor/Claude-3-Opus-Claude-3.5-Sonnnet-9k}.}, \textbf{GPTeacher/GPT-4 General-Instruct}\footnote{\url{https://github.com/teknium1/GPTeacher/tree/main/Instruct}.}, and \textbf{Instruction in the Wild}\footnote{\url{https://github.com/XueFuzhao/InstructionWild}.}.

\end{itemize}

\section{Data Preprocessing for the \Dataset and \Bench Datasets}
\label{sec:data_preprocessing}
\mypara{\Dataset Dataset}
For the \Dataset dataset, we exclude texts with fewer than $150$ characters (including spaces) and texts where the proportion of English content is below $90\%$. Plus, we observe that LLMs' responses often contain redundant or irrelevant content. For example, many LLMs' generated texts include irrelevant phrases at the beginning, such as \highlight{``Of course…''} or \highlight{``Hey there…''}. Additionally, we find that responses generated by the Llama model often repetitively display strings of numbers or specific symbols, hitting the generation length limit instead of providing a complete answer. like \highlight{``….throwaway11111…''}. We filter and remove these anomalous generated contents to enhance the accuracy of our dataset.

\mypara{\Bench Dataset}
For the \Bench dataset, we exclude texts with fewer than 150 characters (including spaces) and texts where the proportion of English content is below $90\%$.

\section{Task Prompts for Generated AIGTs from Social Media}
\label{sec:task_prompt}

Inspired by~\cite{liu2023detectability}, below are designed task prompts for polishing texts on Medium, Quora, and Reddit.
\begin{mdframed}[nobreak=true, backgroundcolor=blue!10, linecolor=blue!50!black, linewidth=2pt, roundcorner=10pt]
Please act as a social media platform {Medium/Quora/Reddit} content creator. 

Your task is to polish the following content. Follow these guidelines:

1. Ensure the content flows naturally and is enjoyable to read.

2. Use simple and relatable language to connect with a broad audience.

3. Highlight key points in a concise and impactful way.

4. Make the content feel more conversational and friendly.

5. Where appropriate, add an engaging tone to draw the reader in.

6. Respond with the revised content only and nothing else:

Here is the original content:
``\{content\}''

\end{mdframed}

Below are designed task prompts for answering the questions on Quora and Reddit.
\begin{mdframed}[nobreak=true, backgroundcolor=yellow!10, linecolor=yellow!50!black, linewidth=2pt, roundcorner=10pt]
You are a content creator on {Quora/Reddit}.

Your task is to generate a thoughtful and insightful answer to the following question. Follow these guidelines:

1. Provide a clear and comprehensive explanation that addresses the question thoroughly.

2. Use simple, relatable language to connect with a broad audience, making the content easy to understand.

3. Highlight key points with examples or anecdotes where applicable, to make the answer more engaging.

4. Add a conversational and friendly tone to make the answer feel more approachable.

5. Ensure the answer is well-structured, with an introduction, body, and conclusion, for better readability.

6. Where relevant, include unique insights or perspectives to make the answer stand out.

7. Respond with the generated answer only and nothing else.

Here is the question:
``\{question\}''
\end{mdframed}

Below are two task prompts designed for summarizing Medium articles and writing detailed articles based on those summaries for Medium articles.
\begin{mdframed}[nobreak=true, backgroundcolor=green!10, linecolor=green!50!black, linewidth=2pt, roundcorner=10pt]
You are a helpful, respectful, and honest assistant. 

Summarize the following content succinctly:

``\{content\}''

Summary:

\end{mdframed}

\begin{mdframed}[nobreak=true, backgroundcolor=orange!10, linecolor=orange!50!black, linewidth=2pt, roundcorner=10pt]
You are a helpful, respectful and honest assistant. Always answer as helpfully as possible, while being safe.

Write a detailed article based on the summary below, following these guidelines:

1. Ensure it flows naturally and is enjoyable to read.

2. Use simple and relatable language for a broad audience.

3. Highlight key points in a concise, impactful way.

4. Make it conversational and friendly.

5. Add an engaging tone where appropriate.

Summary:

``\{summary content\}''

Article:
\end{mdframed}

\section{Detailed Performance of \Detector}
\label{sec:detailed_performance}
\Cref{tab:aigtbench_results} presents the performance of \Detector on individual platform-specific datasets within \Bench. 
The results show that \Detector achieves consistently high accuracy across all three platforms, with accuracy scores of $0.995$, $0.999$, and $0.984$ on Medium, Quora, and Reddit, respectively.

\begin{figure}[ht!]
    \centering
    \scalebox{0.8}{
    \includegraphics[width=\linewidth]{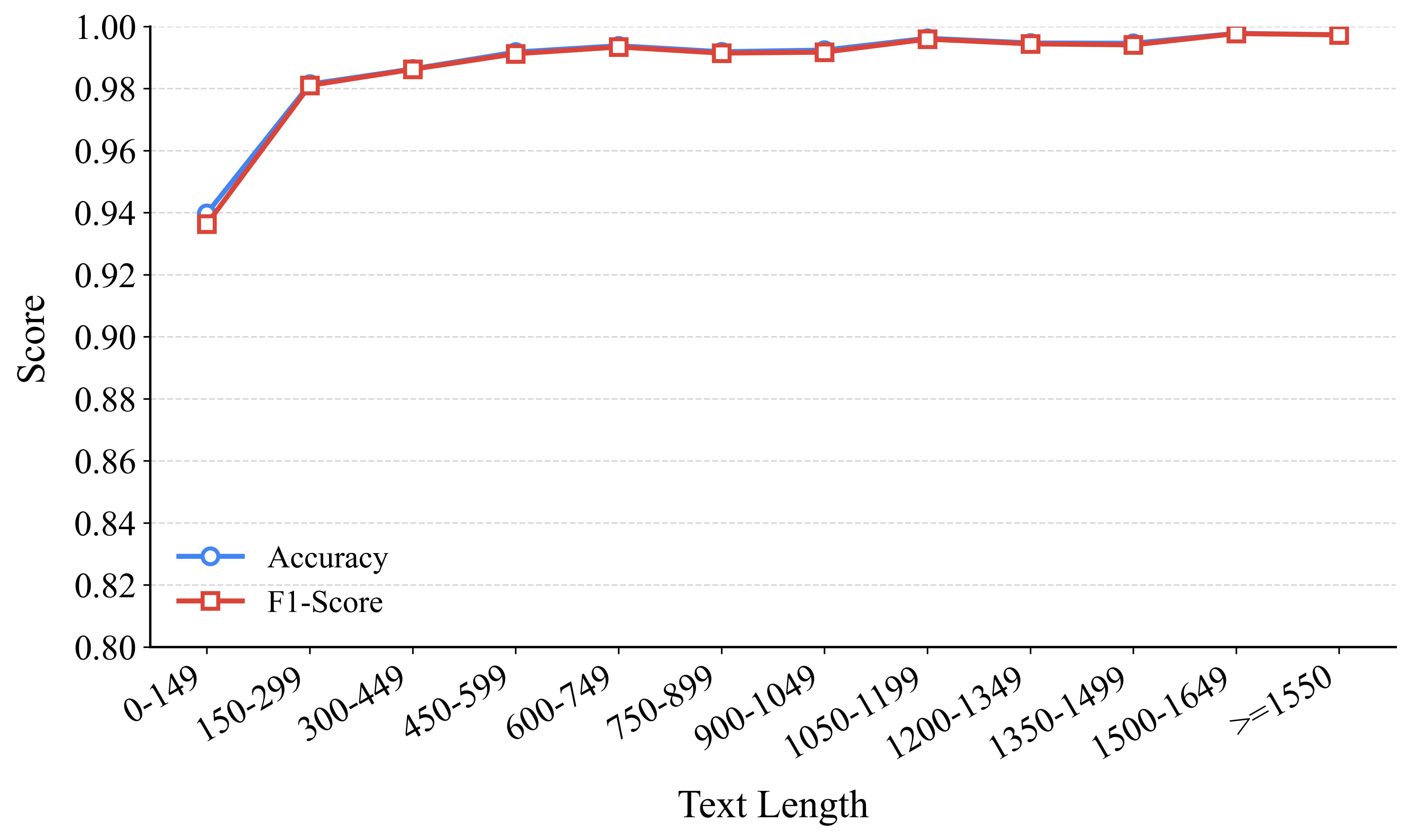}
    }
    \caption{Performance of \Detector across varying text lengths on \Bench.}
    \label{fig:performance_on_different_text_lengths}
\end{figure}

\Cref{fig:performance_on_different_text_lengths} illustrates the accuracy and F1-score across different text lengths in \Bench. 
We observe that accuracy is relatively lower for shorter texts, with an accuracy of approximately $0.940$ for texts between $0-149$ characters. 
However, for texts exceeding $150$ characters, accuracy improves significantly to above $0.980$. 
Both accuracy and F1-score continue to increase as text length increases.

To ensure the reliability of our conclusions, we filter out texts shorter than $150$ characters from \Dataset in our paper.

\section{Details About the Collection of High-Frequency Words and Model Interpretation Analysis Methods}

\subsection{Collection of High-Frequency Words}
\label{sec:collection-high-frequency-wrods}

We use the Spacy library~\cite{Honnibal_spaCy_Industrial-strength_Natural_2020} to classify the part-of-speech of words in the \Bench, specifically dividing them into adjectives, adverbs, and connectives. We then select around the top 20 words for human-preferred and AI-preferred categories, respectively. For detailed results, refer to~\Cref{tab:words_categories}.

\subsection{Model Interpretation Analyze Methods}
\label{sec:model-interpretation-analyze-methods}

Here are the details and how we implement the two different methods:

\begin{itemize}
    \item \textbf{Integrated Gradients} give an importance score to each input value by calculating the gradient of the detector. We follow~\cite{kokhlikyan2020captum} for implementation.
    \item \textbf{Shapley Value} is originally introduced in~\cite{shapley1953value} and recently apply to machine learning interpretation. It quantifies the impact of each feature by perturbing the input value and observing the contributions in the prediction. We follow~\cite{scott2017unified} for implementation.
\end{itemize}

\section{Deeper Analysis of the Result}
\label{sec:deeper-analysis}

Regarding \textbf{the results of ``Evaluation on Social Media Platforms''}, the AAR observed on Medium, Quora, and Reddit show divergent trends in AI adoption. 
Medium's consistently high AAR suggests that its longer-form, polished content format may be particularly conducive to AI-assisted creation. 
The early and substantial adoption may indicate that Medium's creators see AI tools as valuable for improving article efficiency.
While Medium's policy permits AI-assisted content with disclosure, our analysis (based on syntactic patterns and keyword detection) finds that just 0.81\% of authors compiled. 
This suggests that many users may not be following the platform's guidelines.
On the other hand, Quora's adoption pattern tends to fluctuate. 
The platform experienced a temporary surge after introducing the Poe (AI tools), but this trend did not persist, possibly indicating a misalignment between AI capabilities and community expectations.
In contrast, Reddit maintains a low AAR, strongly resisting AIGT despite the platform's text-centric nature. 
This phenomenon may stem from Reddit's vertically organized subcommunities (subreddits), where members reinforce identity through shared jargon and memes, content that current LLMs struggle to generate. 
On top of that, active community moderation behaviors serve as a self-purifying mechanism, such as downvoting "overly fluent" or suspicious content. Thus, the platform's strong community governance and inherent culture pose barriers to AI content adoption.

Regarding \textbf{the underlying reasons for ``significantly higher AAR in technical fields (including Technology and Software Development)''}, One possible explanation is that technical content tends to be highly structured, which fits well with the generation patterns of LLMs. 
Another factor is that technical terminology inherently possesses objectivity, with readers typically focusing more on information accuracy rather than storytelling, making AIGTs easier to accept.
Moreover, technical communities tend to view AI tools as ``advanced productivity enhancers'' rather than ethical threats, further normalizing the use of AIGTs in technical subjects.

The phenomenon of \textbf{``AIGTs receiving lower engagement than HWTs but with limited disparity''} may be related to cognitive psychology factors. 
We speculate that some users possibly rely on heuristic judgments (such as language fluency and information density) to evaluate content, making it difficult for them to clearly distinguish between high-quality AIGTs and HWTs. 
This explains why some AIGT still achieves fundamental interaction measures. However, when content involves subjective perspectives, users tend to activate deeper analytical processes, and AIGTs often lack personal experiential grounding. 
This flattening of affective expression triggers user vigilance, consequently diminishing interaction behaviors.

When it comes to \textbf{why ``low-follower authors tend to rely more on AI''}, one likely explanation is that some authors with smaller followers (0-1K) are often less experienced in writing quality content, leading them to prioritize using AI to sustain their content output. 
Meanwhile, high-follower authors (>5K) often consciously limit their use of AI. 
By doing so, they can preserve their authentic human writing style, which helps reinforce their expert credibility and protects their relationship with their audience.

We have the following recommendations for platform operators, content creators, and readers:
\begin{itemize}
    \item \textbf{For platform operators:} AI-generated content presents dual challenges of content authenticity and user trust. 
    Our research shows that although platforms have established AI content disclosure policies, compliance rates fall far below expectations. 
    Platforms may choose to implement defensive strategies, such as AI detection and content screening, or integrative strategies incorporating AI as a creative aid. 
    Looking ahead, we foresee that AI content labeling will become standardized, alliances will form among platforms to harmonize policies, and the relationship between AI and human creation will evolve from substitution to symbiosis. 
    This requires platforms to develop mechanisms that foster such an ecosystem proactively.
    \item \textbf{For content creators:} We recommend establishing clear guidelines for AI usage: AI tools should serve as assistants, primarily supporting essential tasks such as grammar checks and formatting adjustments.
    The core aspects of creation, central arguments, narrative logic, and emotional expression, must remain human-driven.
    Besides, if human-original content is mistakenly flagged as AIGT, we encourage authors to appeal and request formal disclosure of the decision basis. 
    This demand for transparency safeguards creators' rights and drives platforms to refine their AIGT detection algorithms, leading to fairer content evaluation mechanisms.
    \item \textbf{For readers:} We recommend cultivating critical thinking when engaging with AI content. 
    For instance, check for AI assistance disclosures at the beginning of articles, analyze texts to learn general AI-generated features, and maintain skepticism toward factual claims.
    AI may provide incorrect answers to complex questions. Furthermore, when encountering uncertain content, readers can use accessible tools (e.g., GPTZero) to aid judgment.
    We also urge readers to report unlabeled AI content to platform administrators, helping maintain the community's integrity.
\end{itemize}

\begin{table}[H]
    \centering
    \resizebox{0.6\linewidth}{!}{
        \begin{tabular}{@{} c c c @{}}
            \toprule
            \textbf{Dataset} & \textbf{Type} & \textbf{Sentence Number} \\
            \midrule
            \multirow{3}{*}{Medium} & Llama Series  & $1,881,733$ \\
                   & GPT Series   & $681,480$ \\
                   & Human        & $2,033,105$ \\
            \midrule
            \multirow{3}{*}{Quora}  & Llama Series  & $1,974,368$ \\
                   & GPT Series   & $721,878$  \\
                   & Human        & $569,749$ \\
            \midrule
            \multirow{3}{*}{Reddit} & Llama Series  & $2,892,584$ \\
                   & GPT Series   & $1,391,054$ \\
                   & Human        & $2,695,271$ \\
            \midrule
            \multirow{2}{*}{Total}  & AIGTs       & $9,543,097$ \\
                   & HWTs       & $5,298,125$ \\
            \bottomrule
        \end{tabular}
    }
    \caption{Sentence number statistics of our generated datasets (Llama Series include Llama-1, 2, 3; GPT Series include GPT-3.5, GPT4o-mini).}
    \label{tab:generated}
\end{table}

\section{Ablation Study on Social Media Data}
\label{sec:ablation_study}
To validate the necessity of the social media data collection for \Bench, we conducted an ablation study comparing \Detector performance with and without the social media training subsets on \Bench. As shown in Table~\ref{tab:ablation_results}, \Detector trained only on open-source datasets exhibits poor performance when evaluated on social media test sets, particularly on Quora and Reddit platforms. This performance gap demonstrates that traditional benchmarks fail to capture the linguistic patterns and stylistic variations inherent in social media content. The new collected datasets created address this issue, improving model robustness.

\begin{table}[htbp]
\centering
\caption{Performance comparison of \Detector trained on \Bench without social media data.}
\label{tab:ablation_results}
\begin{tabular}{lcc}
\toprule
\textbf{Platform} & \textbf{Accuracy} & \textbf{F1-score} \\
\midrule
Medium     & 0.975 & 0.967 \\
Quora      & 0.684 & 0.678 \\
Reddit     & 0.631 & 0.608 \\
\bottomrule
\end{tabular}
\vspace{0.2cm}
\end{table}

\begin{figure}[h!]
    \centering

            \begin{subfigure}{\linewidth}
                \centering
                \scalebox{0.9}{%
                \includegraphics[width=\linewidth]{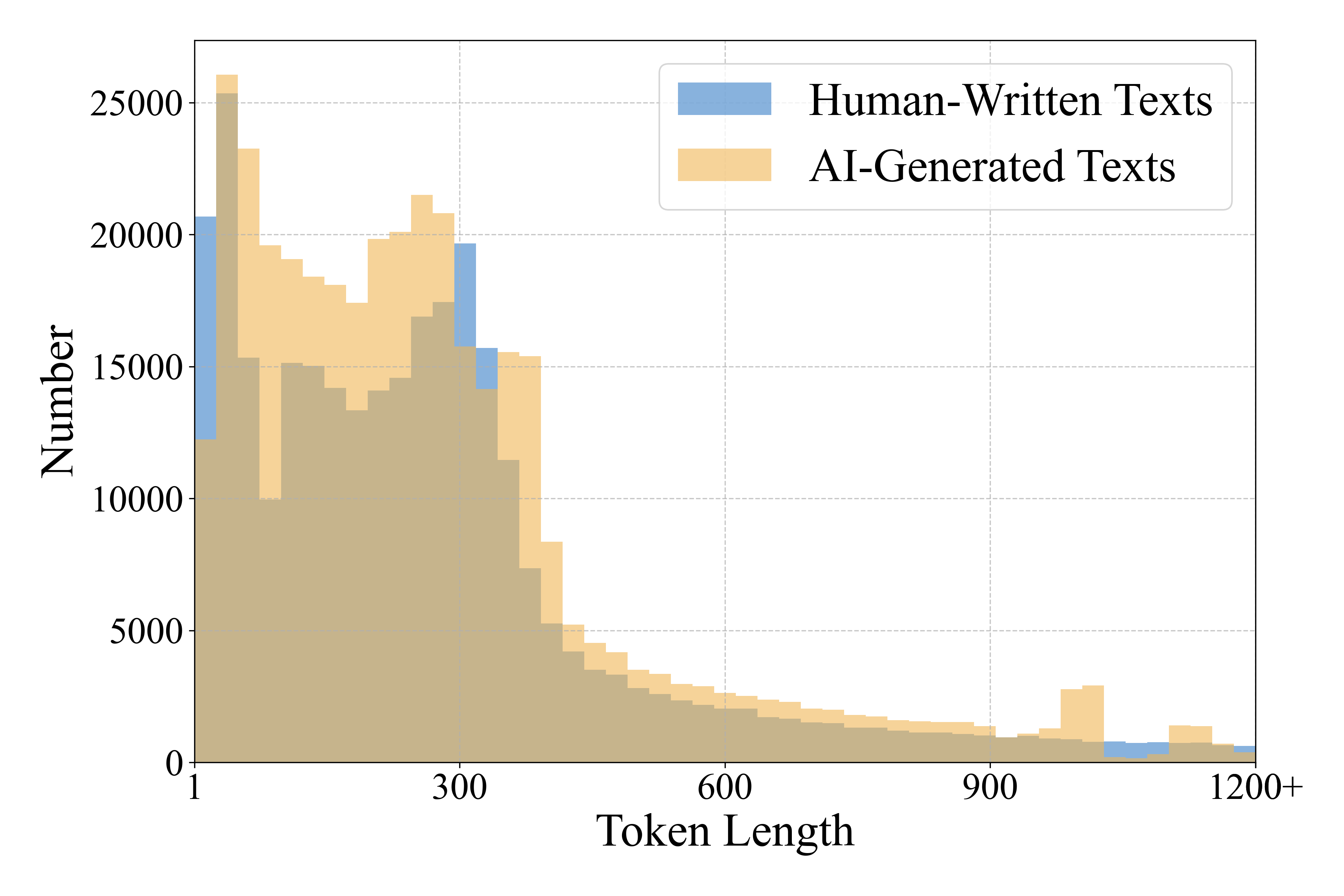}}
                \caption{Token length distribution in the training set.}
            \end{subfigure}
            \vskip 1em
            \begin{subfigure}{\linewidth}
                \centering
                \scalebox{0.9}{
                \includegraphics[width=\linewidth]{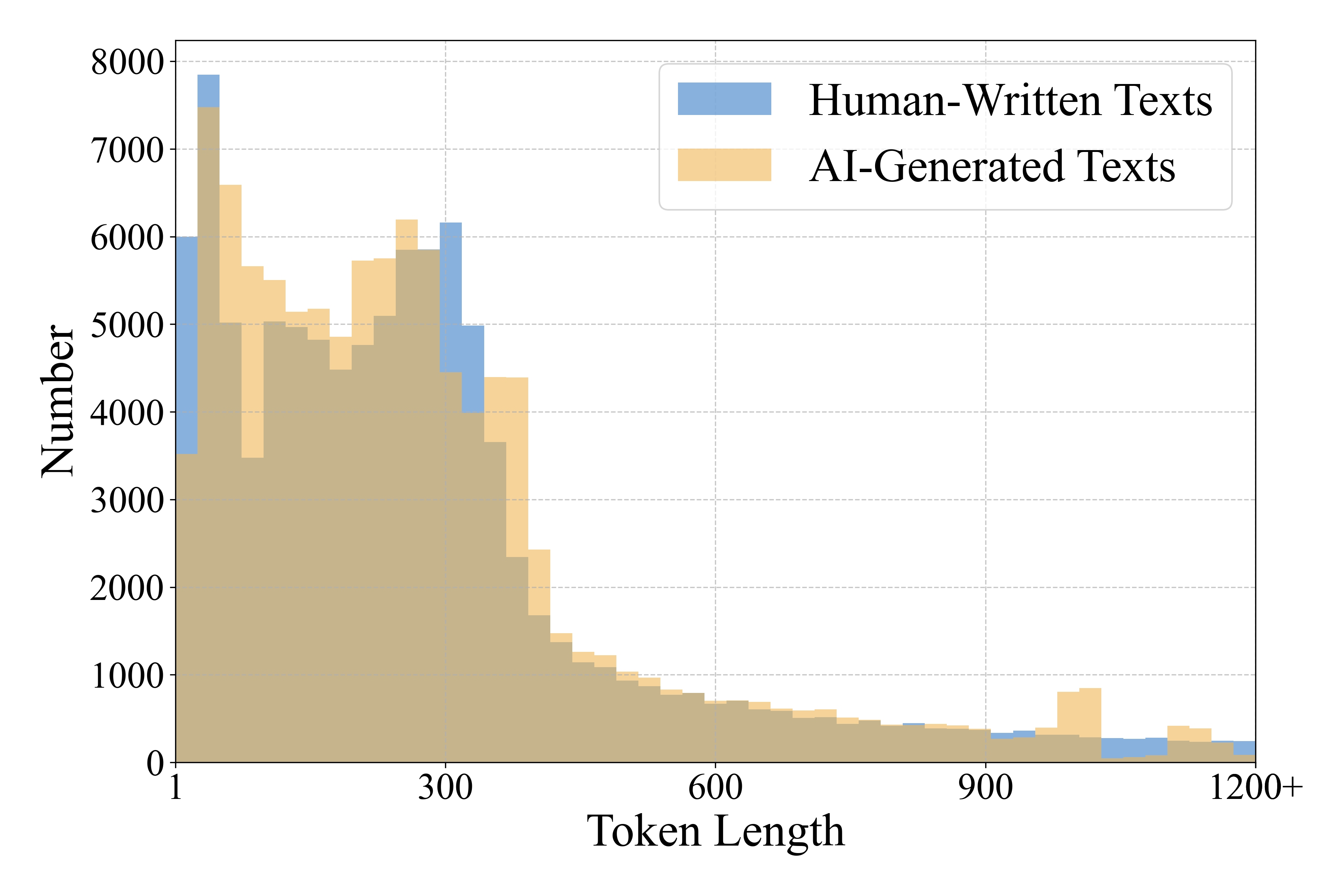}}
                \caption{Token length distribution in the testing set.}
            \end{subfigure}
            \vskip 1em
            \begin{subfigure}{\linewidth}
                \centering
                \scalebox{0.9}{
                \includegraphics[width=\linewidth]{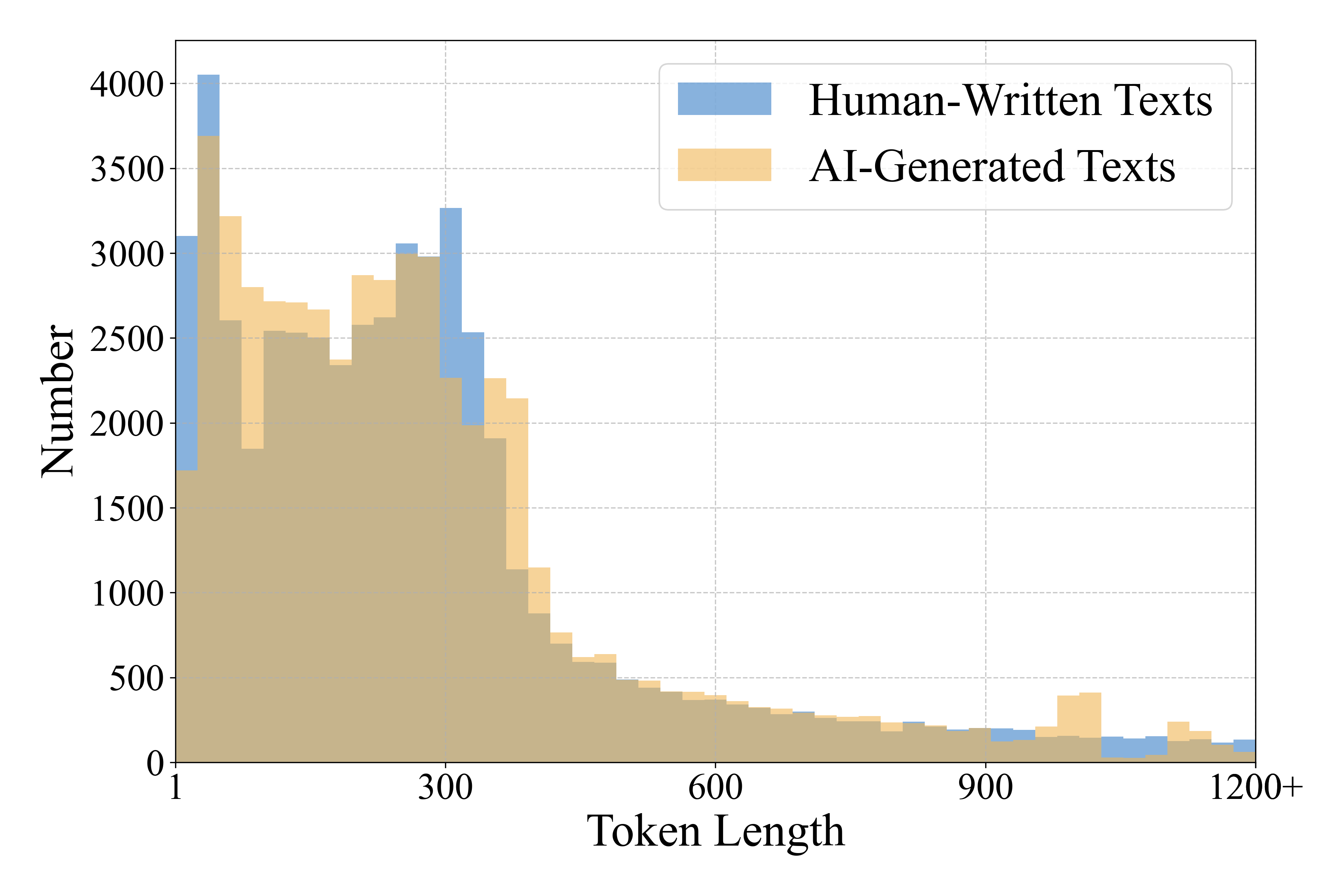}}
                \caption{Token length distribution in the validation set.}
            \end{subfigure}
            \caption{Token length distribution in the training, testing, and validation sets, calculated by the Llama-2 tokenizer~\cite{touvron2023llama2}.}
            \label{fig:train_token_length_distribution}
\end{figure}

\begin{figure}[h!]
    \centering
    \includegraphics[width=\linewidth]{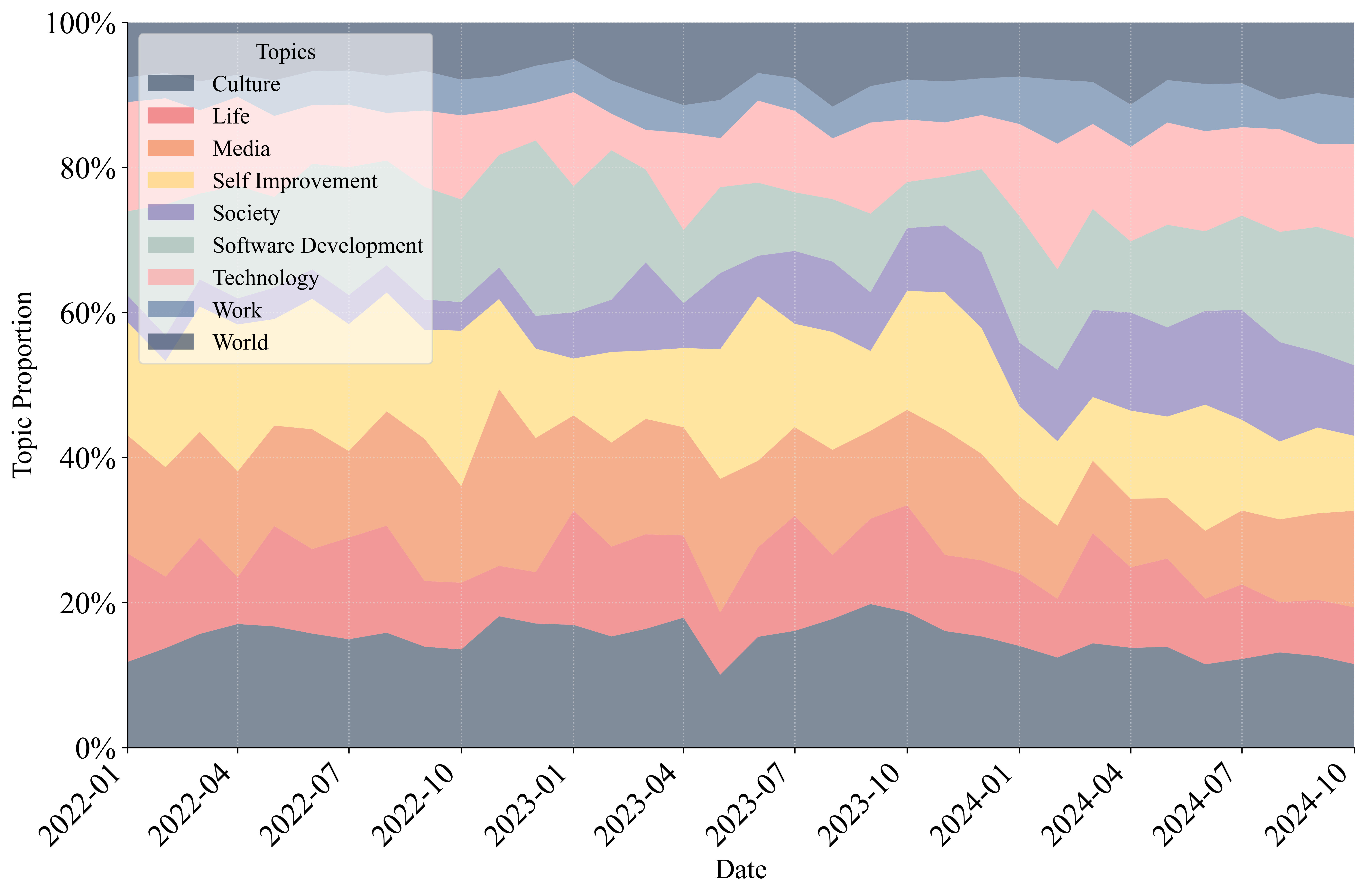}
    \caption{Stacked area chart shows the monthly proportions of $9$ topics.}
    \label{fig:stacked-area-chart}
\end{figure}

\begin{figure}[h!]
    \centering
    \begin{subfigure}[b]{0.46\linewidth}
        \centering
        \includegraphics[width=\textwidth]{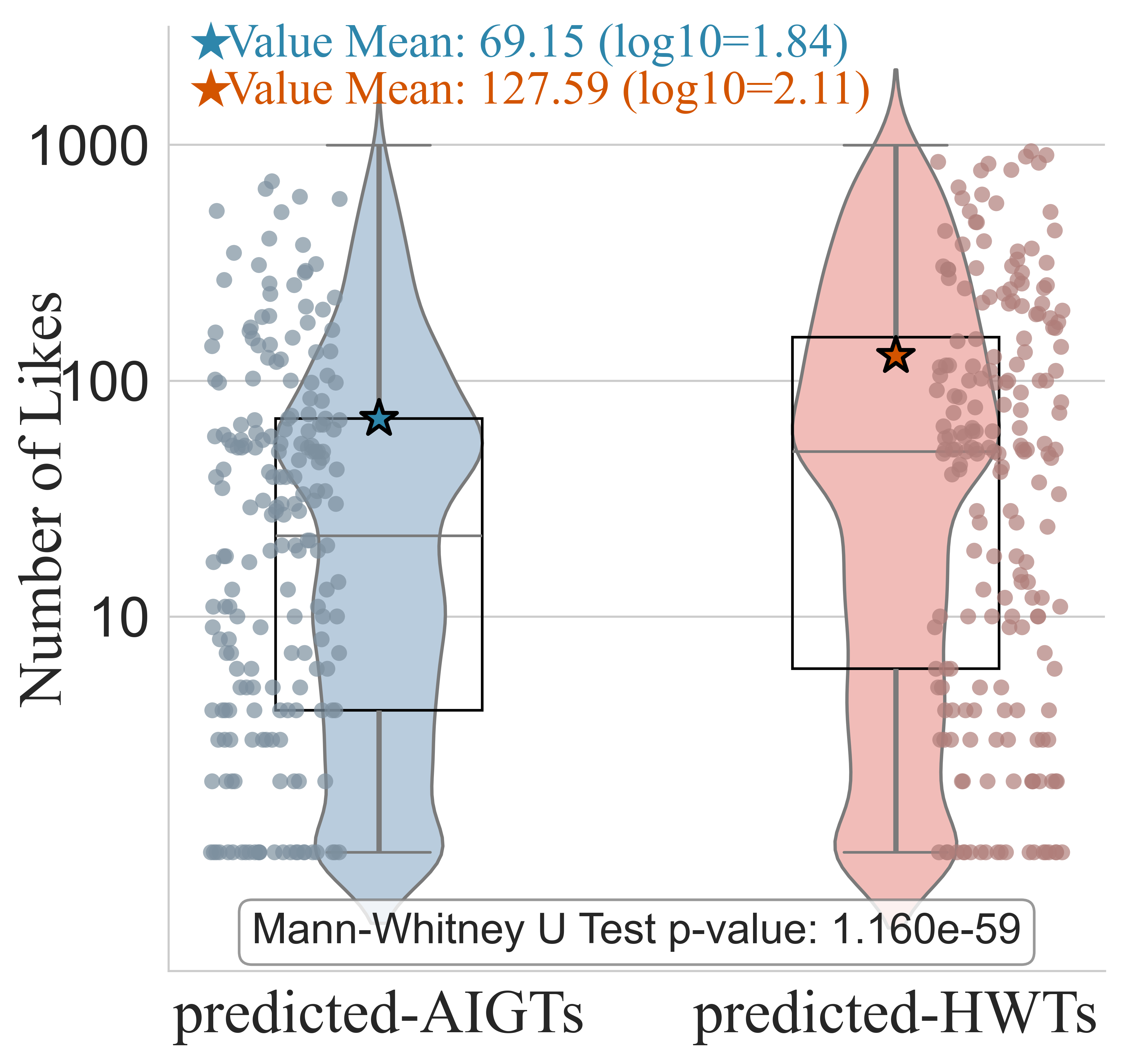}
        \caption{Number of Likes.}
        \label{fig:claps}
    \end{subfigure}
    \hfill
    \begin{subfigure}[b]{0.46\linewidth}
        \centering
        \includegraphics[width=\textwidth]{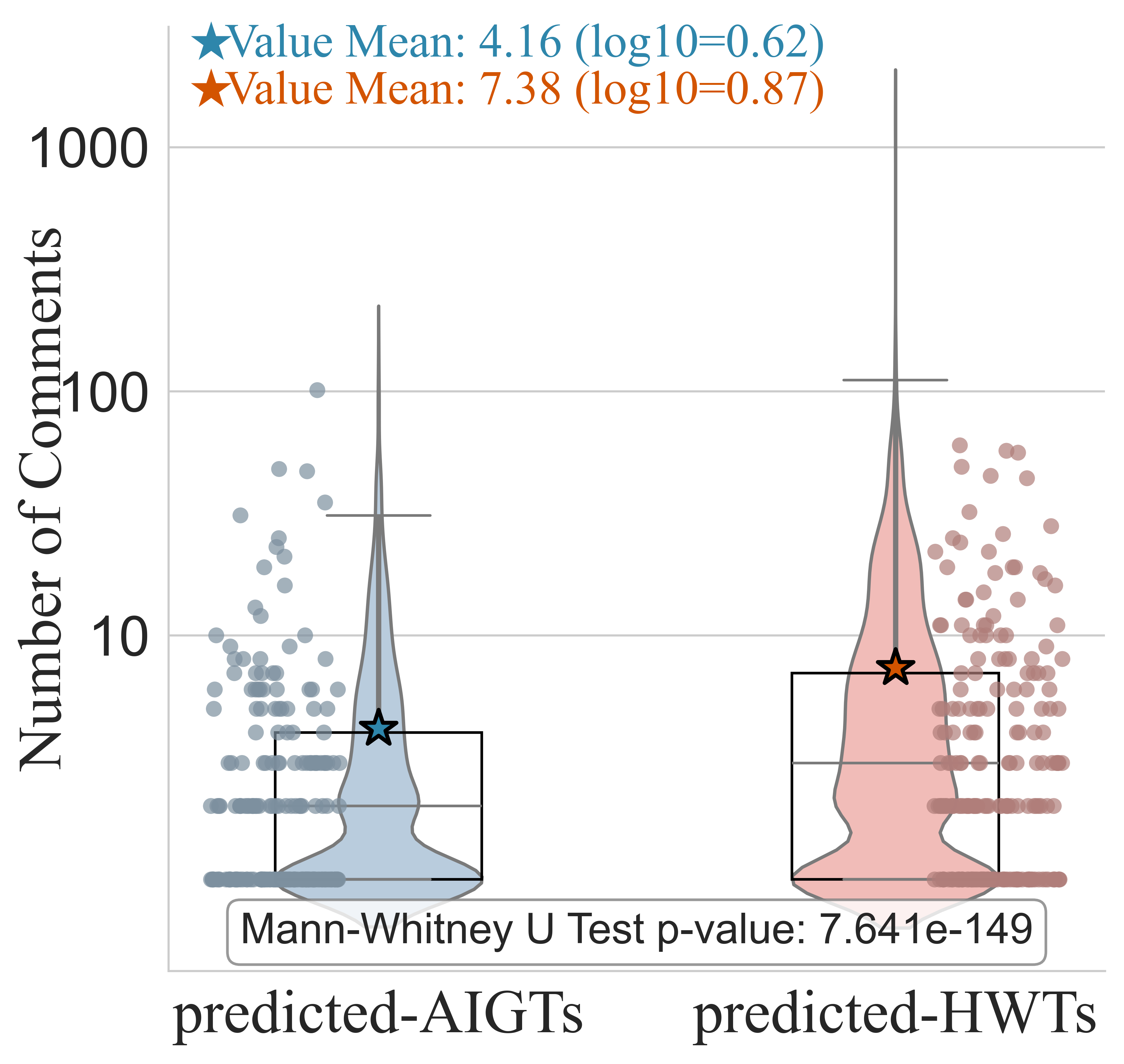}
        \caption{Number of Comments.}
        \label{fig:comments}
    \end{subfigure}
    \caption{Differences between predicted AIGTs and predicted HWTs compressed using a log10 transformation.}
    \label{fig:combined}
\end{figure}

\begin{figure}[h!]
    \centering
    \includegraphics[width=\linewidth]{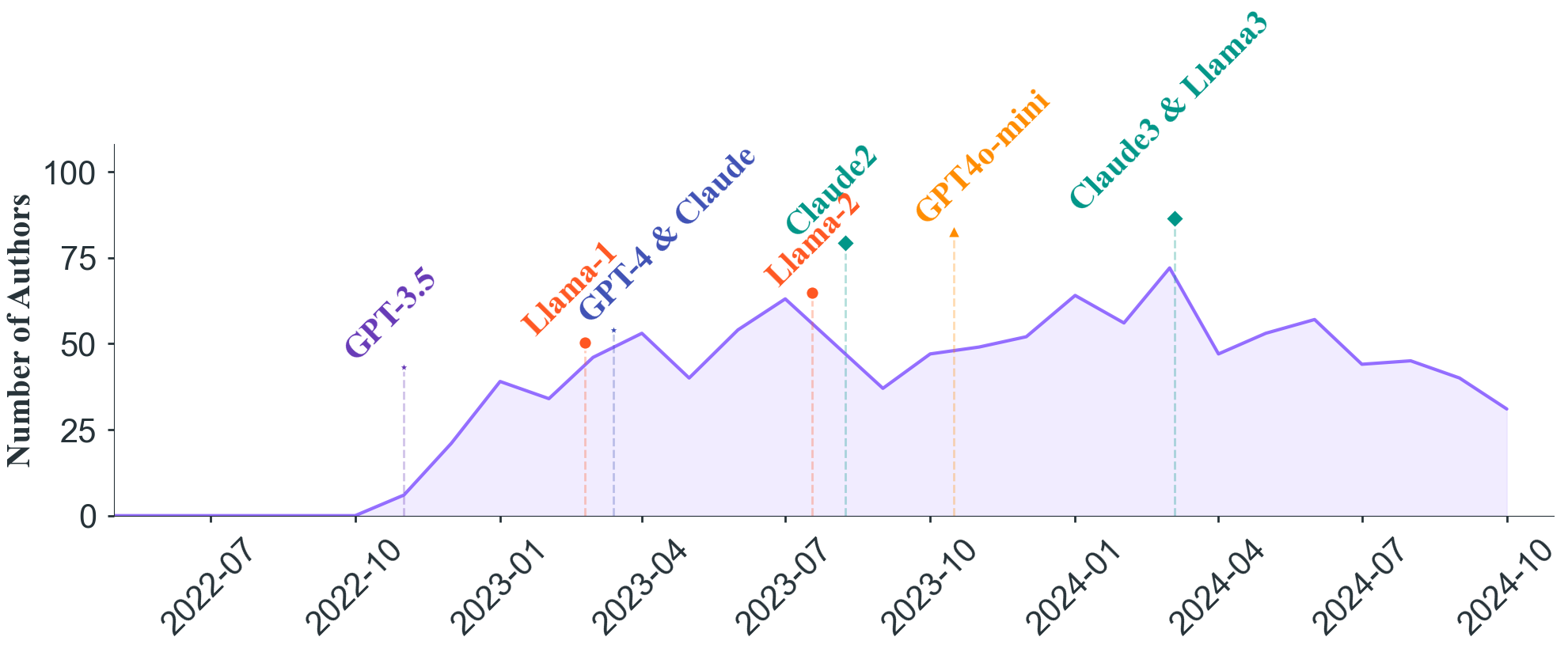}
    \caption{Timeline of authors' earliest adoption of AIGTs.}
    \label{fig:timeline_early_author_aigts}
\end{figure}

\clearpage

\begin{table*}[h!]
\centering
\small
\begin{tabular}{@{\extracolsep{4pt}}cccc@{}}
\toprule
\textbf{Dataset}    & \textbf{Type}          & \textbf{Sentence Number} & \textbf{Domain}\\
\midrule
\multirow{5}{*}{MGT-Academic~\cite{liu2024generalizationabilitymachinegeneratedtext}}  
                                & Llama3               & $1,478,485$  & \multirow{5}{*}{\makecell{STEM (Physics, Math, Biology, CS,\\ EE, Statistics, Chemistry, Medicine),\\ Social Science (Education, Economy, \\ Management), Humanities (Literature, Law,\\ Art, History, Philosophy)}} \\
                                & Mixtral 8$\times$7B & $2,639,498$  \\
                                & Moonshot             & $726,357$  \\
                                & GPT-3.5        & $1,611,244$  \\
                                & Human                & $6,007,476$  \\
\midrule
\multirow{2}{*}{Coco-GPT3.5~\cite{liu2023coco}}    
                                & GPT-3.5     & $79,647$  & \multirow{2}{*}{News} \\
                                & Human                & $55,565$  \\
                                
\midrule
\multirow{2}{*}{GPABench2~\cite{liu2023detectability}}      
                                & GPT-3.5        & $12,648,338$ (Sample) & \multirow{2}{*}{Computer Science, Physics, Social Sciences} \\
                                & Human                & $1,065,860$  \\
\midrule
\multirow{4}{*}{LWD~\cite{soto2024few}}            
                                & Llama2               & $94,732$ & \multirow{4}{*}{Finance, Social Media} \\
                                & GPT-3.5        & $95,443$  \\
                                & GPT-4               & $62,632$  \\
                                & Human                & $106,952$  \\
\midrule
\multirow{6}{*}{AIGT~\cite{shi2024ten}}           
                                & Llama2           & $6,967$  & \multirow{6}{*}{Soical media, News, Academic Writing} \\
                                & Alpaca 7B        & $6,083$  \\
                                & Vicuna 13B       & $7,028$  \\
                                & GPT-3.5    & $8,022$  \\
                                & GPT-4            & $7,156$  \\
                                & Human            & $12,228$  \\
\midrule
\multirow{2}{*}{HC3~\cite{guo2023close}}            
                                & GPT-3.5        & $184,692$ & \multirow{2}{*}{\makecell{Open-domain, Finance, Medicine, \\ Law, and Psychology}} \\
                                & Human                & $347,423$  \\
\bottomrule
\end{tabular}
\caption{Statistics of open-source datasets (part 1).}
\label{tab:datasets_tokens_1}
\end{table*}

\begin{table*}[h!]
\centering
\small
\begin{tabular}{@{\extracolsep{4pt}}cccc@{}}
\toprule
\textbf{Dataset}    & \textbf{Type}          & \textbf{Sentence Number} & \textbf{Domain}\\
\midrule
Claude2-Alpaca                & Claude-2           & $404,051$  & Open-domain\\
\midrule
\multirow{2}{*}{Claude-3-Opus-Claude-3.5-Sonnnet-9k} & Claude-3 & $276,246$  & \multirow{2}{*}{Open-domain}\\
                                & Human            & $37,785$ \\
\midrule
\multirow{2}{*}{GPTeacher/GPT-4 General-Instruct}  & GPT-4             & $74,160$ & \multirow{2}{*}{Open-domain} \\
                                & Human            & $24,465$ \\
\midrule
\multirow{2}{*}{Alpaca\_GPT4}  & GPT-4            & $354,801$ &\multirow{2}{*}{Open-domain} \\
                                & Human            & $22,253$ \\
\midrule
\multirow{1}{*}{Instruction in the Wild} & GPT-3.5 & $300, 424$ & Open-domain\\
\bottomrule
\end{tabular}
\caption{Statistics of open-source datasets (part 2).}
\label{tab:datasets_tokens_2}
\end{table*}

\begin{table*}[ht!]
\centering
\small
\begin{tabular}{@{\extracolsep{4pt}}c>{\raggedright\arraybackslash}p{0.7\textwidth}@{}}
\toprule
\textbf{Category} & \textbf{Words} \\
\midrule
\multicolumn{1}{@{}c}{\parbox{0.2\textwidth}{\centering\textbf{Human top} \\ \textbf{frequency words}}} & 
`little', `small', `last', `able', `bad', `next', `right', `most', `long', `old', `much', `sure', `great', `actually', `again', `probably', `much', `very', `pretty', `already', `since', `against', `yet' \\
\midrule
\multicolumn{1}{@{}c}{\parbox{0.2\textwidth}{\centering\textbf{AI top} \\ \textbf{frequency words}}} & 
`various', `significant', `positive', `complex', `original', `free', `specific', `unique', `crucial', `clear', `human', `personal', `essential', `particularly', `especially', `truly', `instead', `here', `rather', `additionally', `despite', `due to', `following' \\
\bottomrule
\end{tabular}
\caption{Categorization of words into human and AI characteristics.}
\label{tab:words_categories}
\end{table*}

\clearpage

\begin{table}[ht!]
    \centering
    \footnotesize
    \scalebox{1.2}{
    \begin{tabular}{@{}ccc@{}}
    \toprule
    \textbf{Platform}   & \textbf{Accuracy}  & \textbf{F1-score}  \\
    \midrule 
    Medium     & $0.995$  & $0.995$  \\ 
    Quora      & $0.999$  & $0.999$  \\
    Reddit     & $0.984$  & $0.984$  \\
    \bottomrule 
    \end{tabular}
    }
    \caption{Performance of \Detector on AIGTs within \Bench across different platforms.}
    \label{tab:aigtbench_results}
\end{table}

\begin{table}[ht!]
    \centering
    \footnotesize
    \begin{tabular}{llcc}
        \toprule
        \multirow{2}{*}{\textbf{Category}} & \multirow{2}{*}{\textbf{Dataset}} & \multicolumn{2}{c}{\textbf{Performance}} \\
        \cmidrule(lr){3-4}
        & & \textbf{Accuracy} & \textbf{F1-score} \\
        \midrule
        \multirow{4}{*}{\makecell{Unseen\\ Model}}  
        & QwQ-32B-Preview~\cite{magpie_reasoning_v1_150k_cot_qwq}  & 0.999 & 0.999 \\
        & Gemini-2.0-Flash~\cite{customsharegpt}  & 0.993 & 0.997 \\
        & Deepseek-R1-Llama-70B~\cite{magpie_reasoning_v2_250k_cot_deepseek_r1_llama_70b}  & 0.999 & 0.999 \\
        \midrule
        \multirow{3}{*}{\makecell{Unseen \\ Domain}} 
        & Roleplay-English~\cite{odiagenai_roleplay_english} & 0.999 & 0.999 \\
        & Mannerstral-dataset~\cite{heralax_mannerstral_dataset} & 0.943 & 0.968 \\
        & InternVL-SA-1B-Captio~\cite{openGVLab_unternvl_as_1b_caption} & 0.998 & 0.999 \\
        \bottomrule
    \end{tabular}
    \caption{Test \Detector in the wild (all datasets from HuggingFace).}
    \label{tab:test_detector}
\end{table}

\begin{figure*}[hb!]
    \makebox[\textwidth][l]{
        \resizebox{\textwidth}{!}{
            \includegraphics{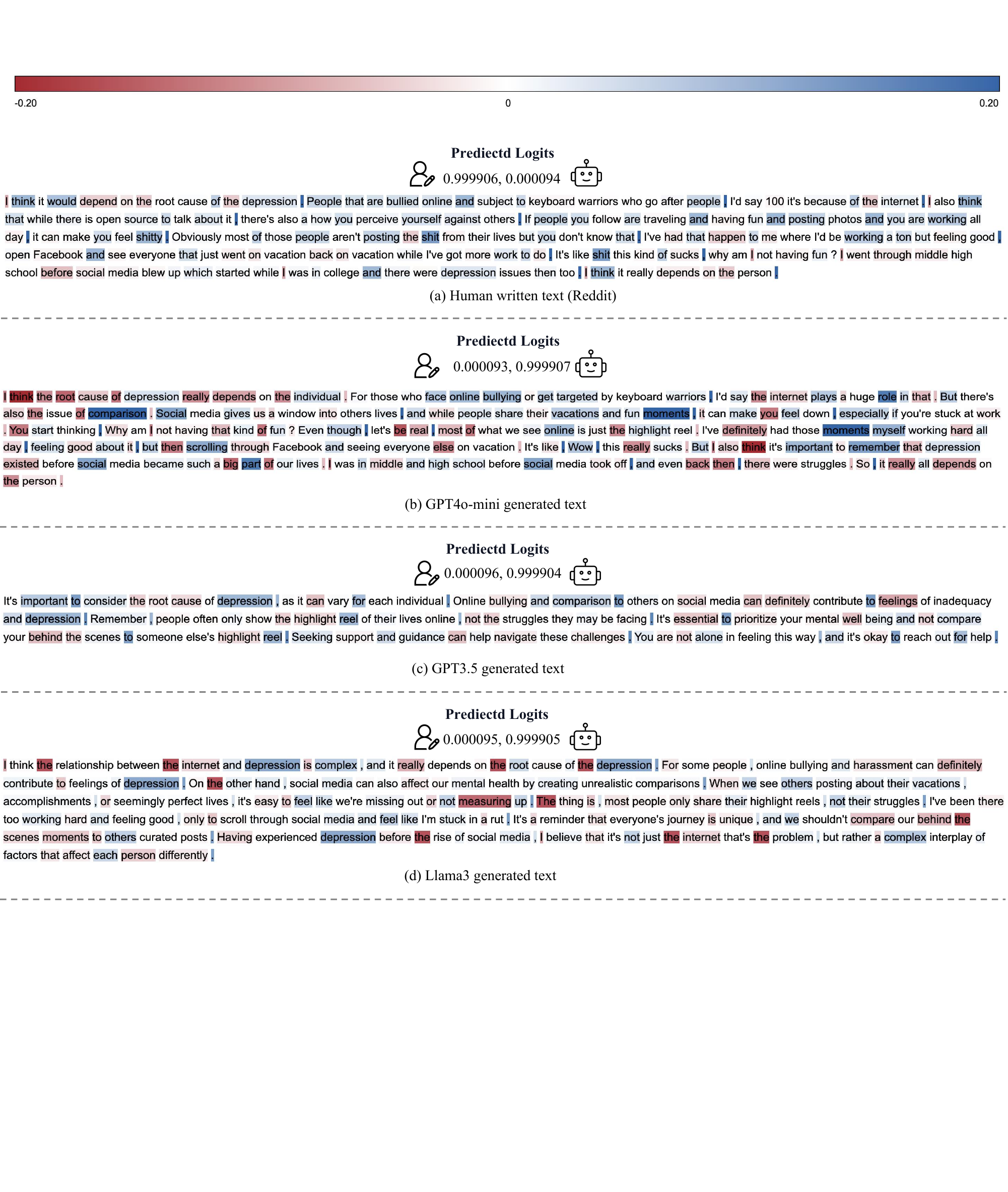}
        }
    }
    \caption{Case study of word-level analysis through Integrated Gradients on Reddit.}
    \label{Reddit_intergrate_grdients}
\end{figure*}

\begin{figure*}[hb!]
    \makebox[\textwidth][l]{
        \resizebox{\textwidth}{!}{            \includegraphics{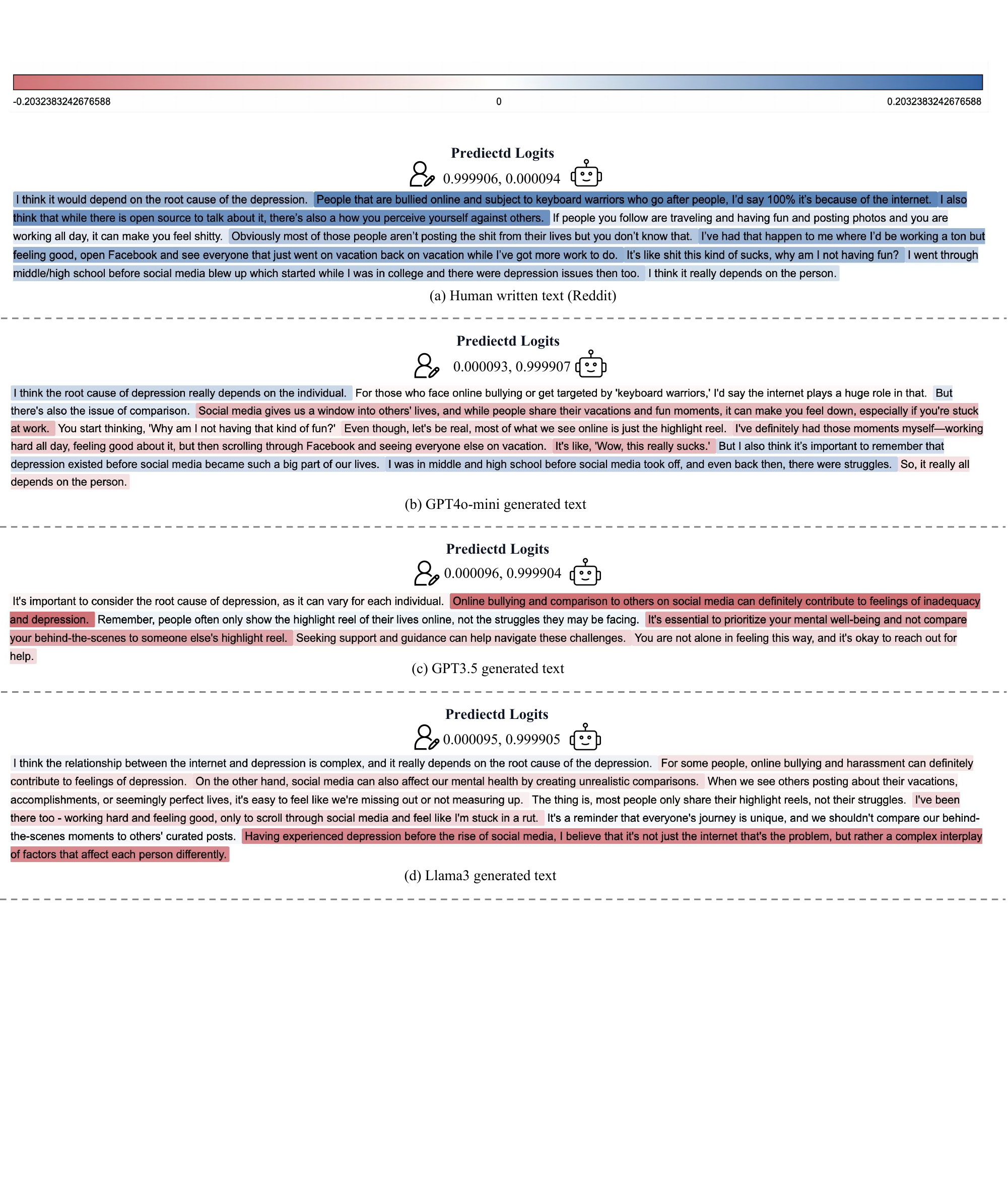}
        }
    }
    \caption{Case study of sentence-level analysis through Shaplay Value on Reddit.}
    \label{Reddit_Sentecet_Level_Shaplay}
\end{figure*}

\begin{figure*}[hb!]
    \makebox[\textwidth][l]{
        \resizebox{\textwidth}{!}{
            \includegraphics{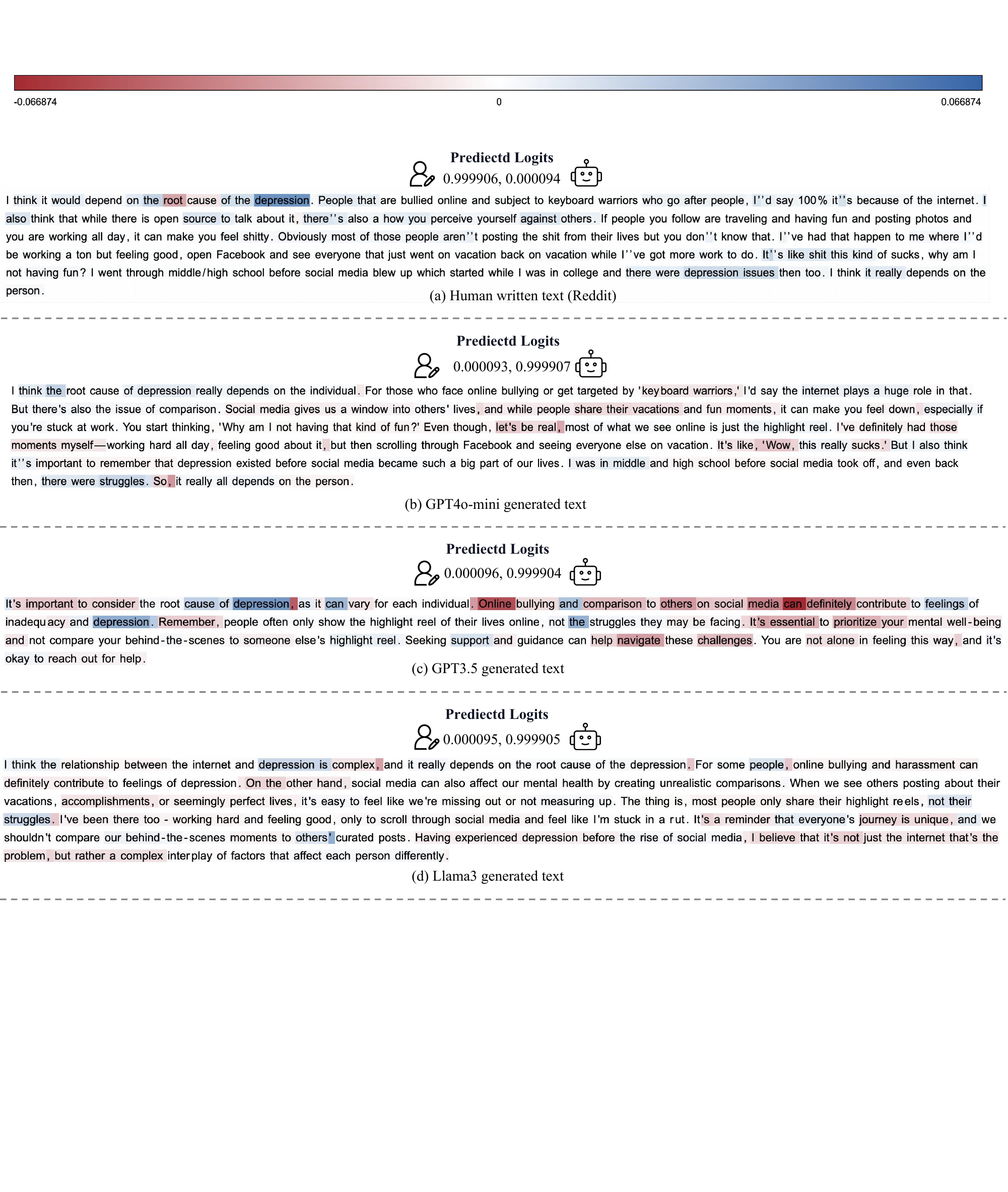}
        }
    }
    \caption{Case study of word-level analysis through Shaplay Value on Reddit.}
    \label{Reddit_word_level_shaplay}
\end{figure*}

\begin{figure*}[hb!]
    \makebox[\textwidth][l]{
        \resizebox{\textwidth}{!}{%
            \includegraphics{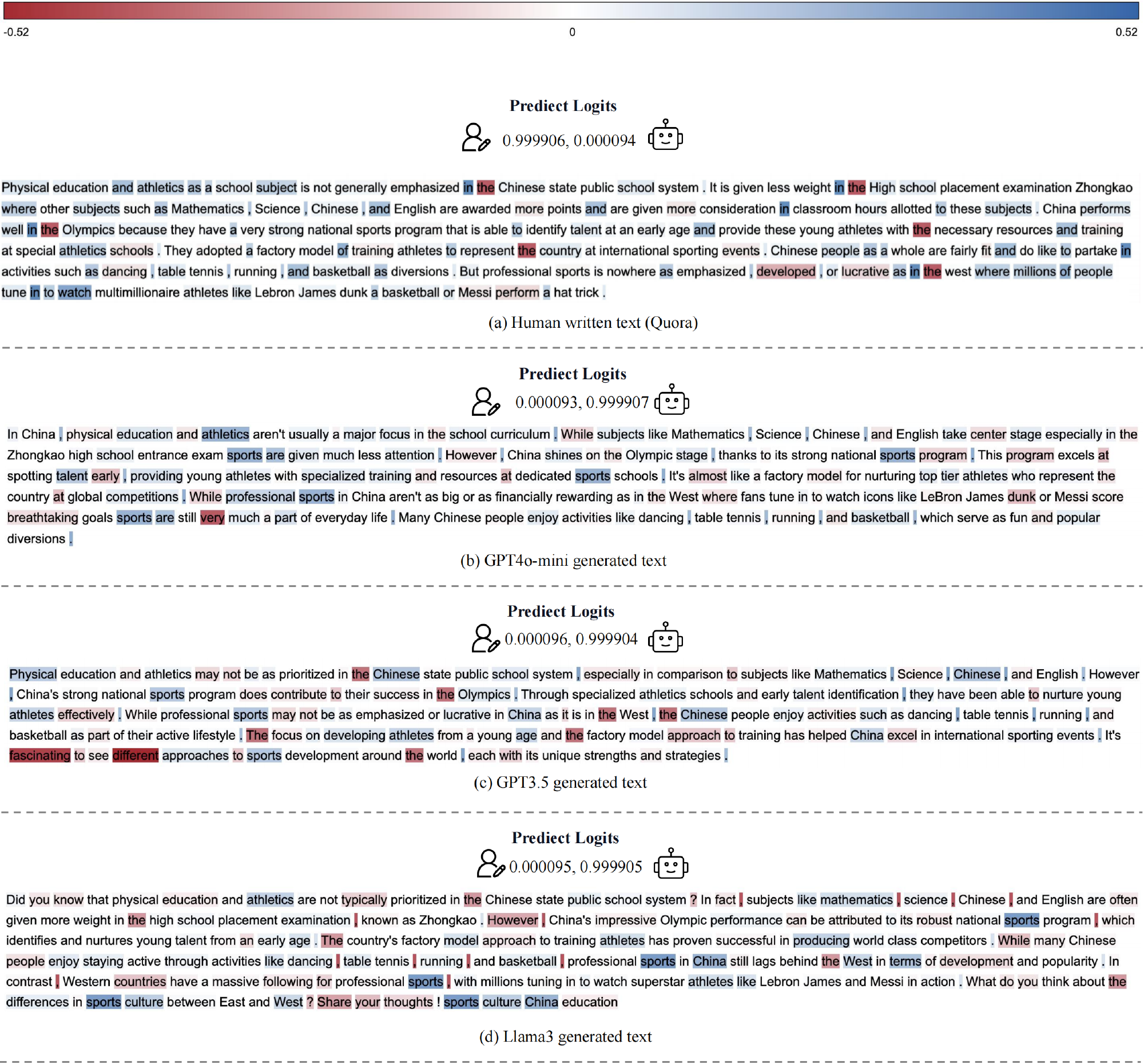}
        }
    }
    \caption{Case study of word-level analysis through Integrated Gradients on Quora.}
    \label{Quora_intergrate_grdients}
\end{figure*}

\begin{figure*}[hb!]
    \makebox[\textwidth][l]{
        \resizebox{\textwidth}{!}{
            \includegraphics{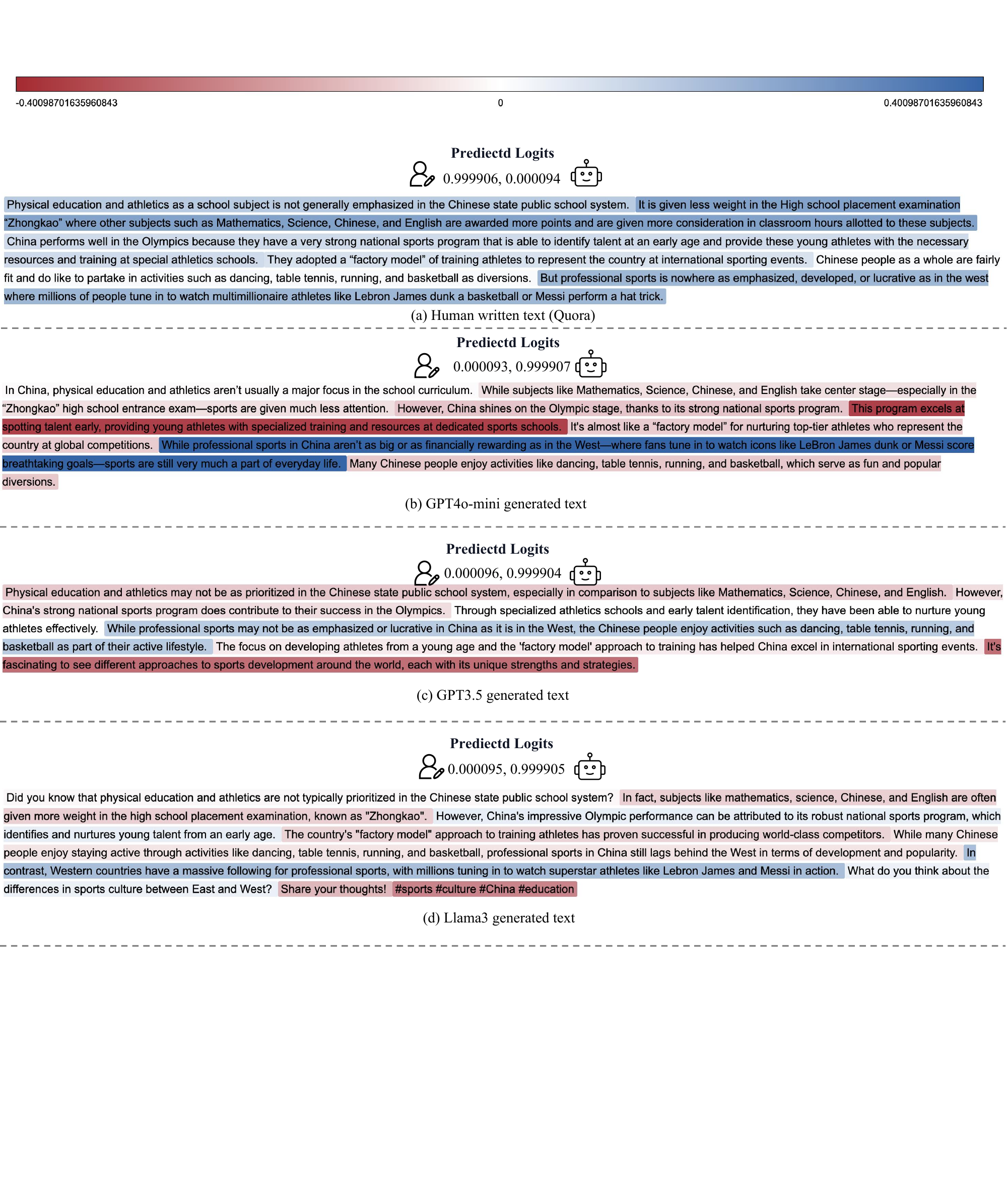}
        }
    }
    \caption{Case study of sentence-level analysis through Shaplay Value on Quora.}
    \label{Quora_sentence_level_shaplay}
\end{figure*}

\begin{figure*}[hb!]
    \makebox[\textwidth][l]{
        \resizebox{\textwidth}{!}{
            \includegraphics{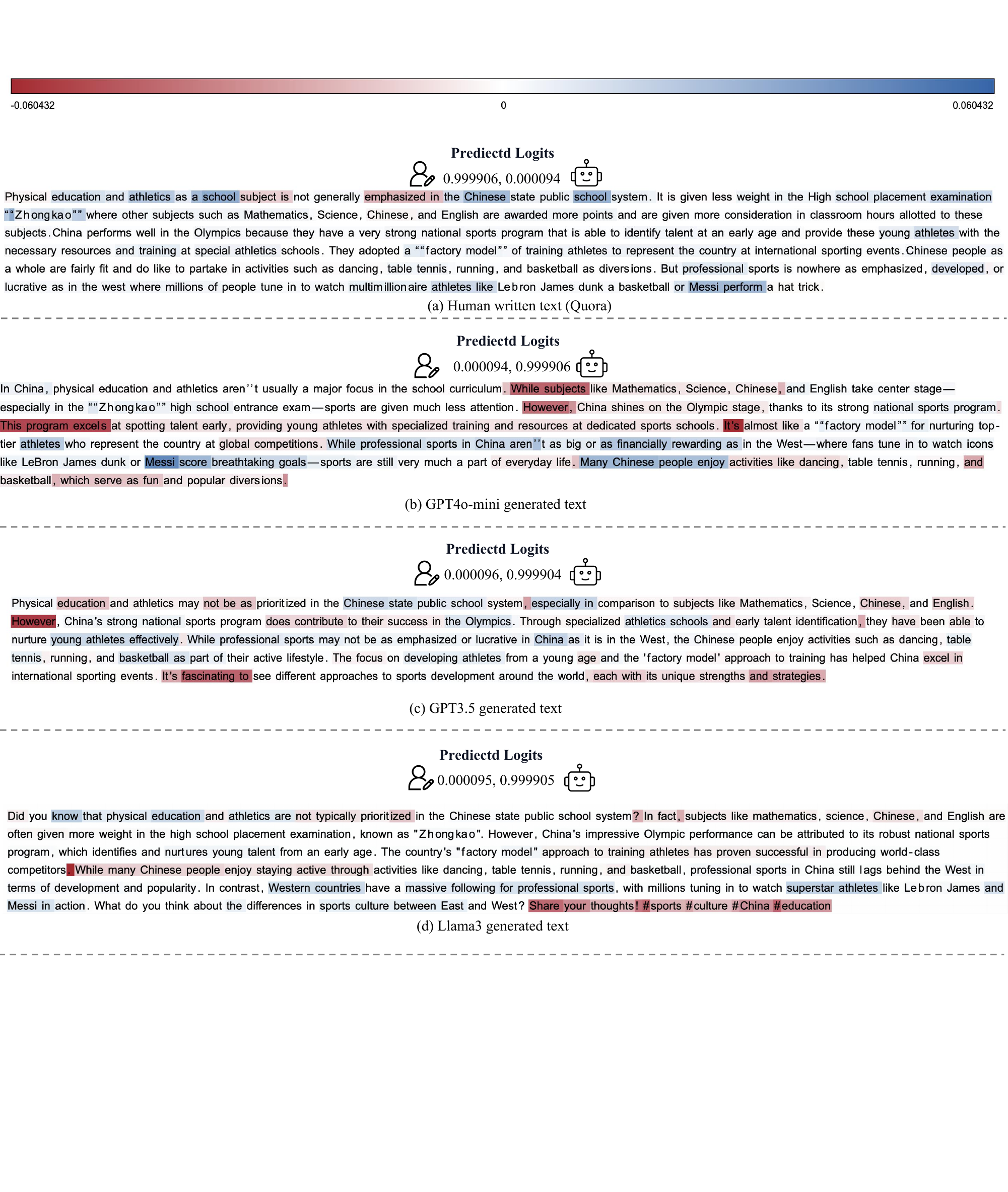}
        }
    }
    \caption{Case study of word-level analysis through Shaplay Value on Quora.}
    \label{Quora_word_level_shaplay}
\end{figure*}

\begin{figure*}[hb!]
    \makebox[\textwidth][l]{
        \resizebox{\textwidth}{!}{%
            \includegraphics{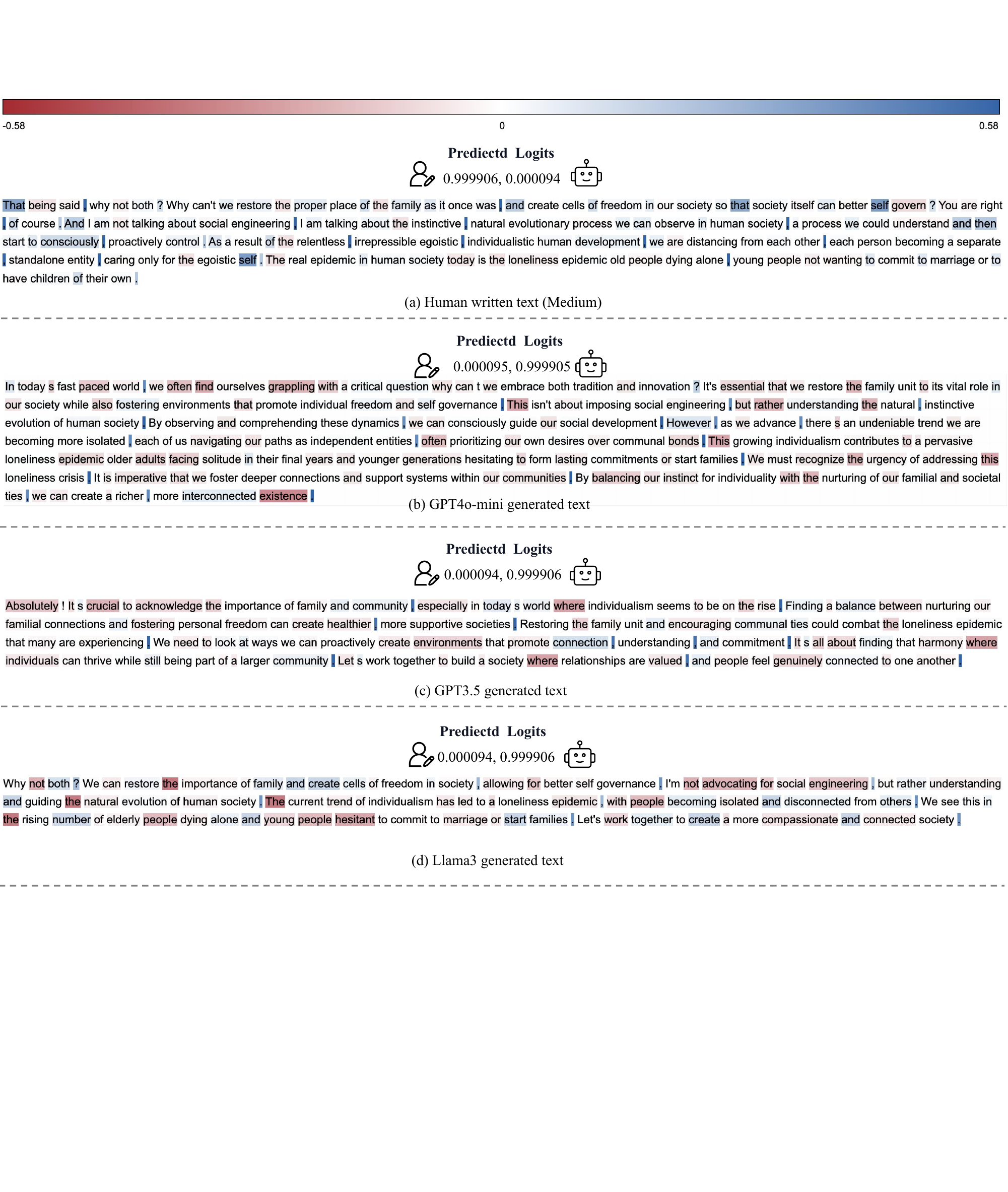}
        }
    }
    \caption{Case study of word-level analysis through Integrated Gradients on Medium.}
    \label{Medium_intergrate_grdients}
\end{figure*}

\begin{figure*}[hb!]
    \makebox[\textwidth][l]{%
        \resizebox{\textwidth}{!}{%
            \includegraphics{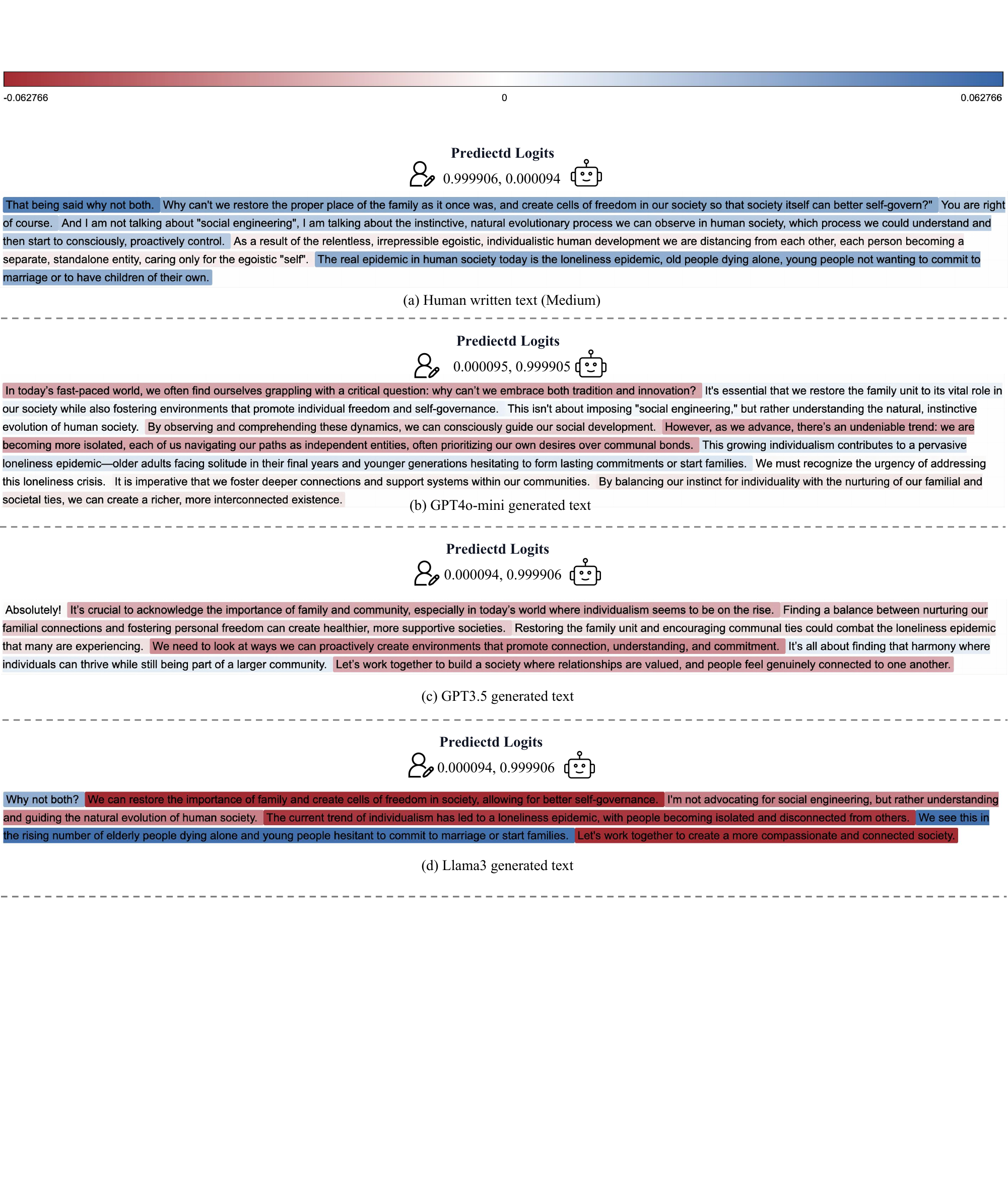}
        }
    }
    \caption{Case study of sentence-level analysis through Shaplay Value on Medium.}
    \label{Medium_sentence_level_shaplay}
\end{figure*}

\begin{figure*}[hb!]
    \makebox[\textwidth][l]{
        \resizebox{\textwidth}{!}{
            \includegraphics{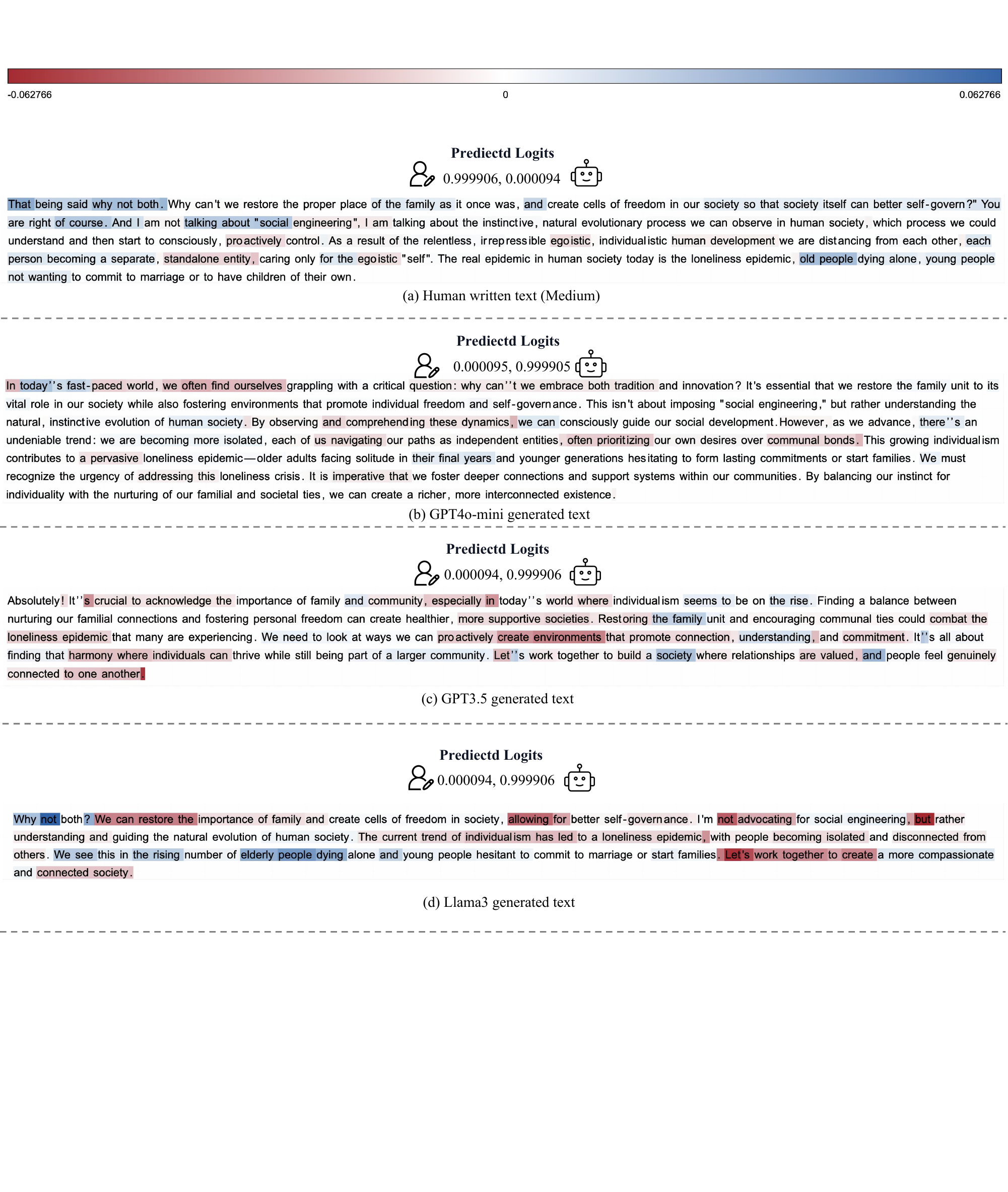}
        }
    }
    \caption{Case study of word-level analysis through Shaplay Value on Medium.}
    \label{Medium_word_level_shaplay}
\end{figure*}

\end{document}